\documentclass[acmsmall,screen,nonacm]{acmart}

\setcopyright{none}
\acmDOI{}
\acmISBN{}
\renewcommand{\footnotetextcopyrightpermission}[1]{}

\makeatletter
\@ifundefined{hyxmp@parse@acmart}{}{\let\hyxmp@parse@acmart\relax}
\makeatother

\usepackage{amsmath}
\usepackage{amsthm}
\usepackage{booktabs}
\usepackage{graphicx}
\usepackage{microtype}
\usepackage{placeins}
\usepackage{tabularx}
\usepackage{xspace}
\usepackage{tikz}
\usetikzlibrary{arrows.meta,positioning,calc}

\definecolor{lawteal}{HTML}{247361}
\definecolor{lawrose}{HTML}{AF4C72}
\definecolor{lawslate}{HTML}{657F9B}

\graphicspath{{figures/}}

\newif\ifresultsready
\resultsreadytrue
\newif\ifforecastresults
\forecastresultsfalse
\newif\ifdurationevidencefinal
\durationevidencefinaltrue
\newcommand{\DOneLevel}{4}
\ifcase\DOneLevel
  \ifresultsready
    \newcommand{\PaperTitle}{Active-Byte Accounting for LLM Decode: Traffic Closure and Energy Non-Closure on GH200}%
  \else
    \newcommand{\PaperTitle}{A Token-Energy Model of LLM Decode: Testing Active Bytes against Joules per Token}%
  \fi
\or
  \newcommand{\PaperTitle}{Active-Byte Accounting: A Conditional HBM-Traffic Bound for LLM Decode}%
\or
  \newcommand{\PaperTitle}{A Two-Component Token-Energy Model of LLM Decode}%
\or
  \newcommand{\PaperTitle}{A Single-Coefficient Active-Byte Law for LLM Decode}%
\or
  \newcommand{\PaperTitle}{The Token-Energy Law of LLM Decode: Active Bytes and Realized Time Account for Joules per Token}%
\else
  \PackageError{active-bytes}{D1 level must be 0, 1, 2, 3, or 4}{}
\fi

\newcommand{\DurationVTwoMedianError}{0.409\%}
\newcommand{\DurationVTwoMaximumError}{2.22\%}

\newcommand{\QFourteenMedianError}{0.130\%}
\newcommand{\QFourteenMaximumError}{1.92\%}

\newcommand{\MistralMedianError}{0.390\%}
\newcommand{\MistralMaximumError}{1.43\%}

\newcommand{\GPUOneTransferMedianError}{2.62\%}
\newcommand{\GPUOneTransferMaximumError}{3.72\%}

\ifresultsready
  \ifforecastresults
    \PackageError{active-bytes}{Disable forecast mode before enabling resultsready}{Measured prose and forecast-watermarked figures cannot coexist in a release build.}
  \fi
  \IfFileExists{results/final_figures_ready.tex}{%
    \newcommand{\FinalFiguresReadyToken}{MEASURED-FIGURES-VERIFIED-V2}
\newcommand{\FinalMeasuredExportDigest}{762a18f9babdab3a16bddfef787e9a66b1de920f1d1348b6a971e1f62e539b83}
\newcommand{\FinalDurationEvidenceDigest}{c1c0fb34cde3c1c697135698b04214c1e71ab307dc1992661c406cfe4ea5ec1f}
\newcommand{\FinalFigureBundleDigest}{9663f36b7ad3b86faf027d5f6c177c93e5943056d70724816ce69102afd2ad87}
    \providecommand{\FinalFiguresReadyToken}{}%
    \providecommand{\FinalMeasuredExportDigest}{}%
    \providecommand{\FinalDurationEvidenceDigest}{}%
    \providecommand{\FinalFigureBundleDigest}{}%
    \ifdefstring{\FinalFiguresReadyToken}{MEASURED-FIGURES-VERIFIED-V2}{\durationevidencefinaltrue}{%
      \PackageError{active-bytes}{Invalid final-figure readiness token}{The final reviewed pipeline must emit MEASURED-FIGURES-VERIFIED-V2.}%
    }%
    \ifdefempty{\FinalMeasuredExportDigest}{%
      \PackageError{active-bytes}{Missing measured-export digest}{The readiness record must bind the figures to the measured export.}%
    }{}%
    \ifdefempty{\FinalDurationEvidenceDigest}{%
      \PackageError{active-bytes}{Missing duration-evidence digest}{The readiness record must bind the V2 figures to the duration-law summary.}%
    }{}%
    \ifdefempty{\FinalFigureBundleDigest}{%
      \PackageError{active-bytes}{Missing final-figure bundle digest}{The readiness record must hash every final figure.}%
    }{}%
  }{%
    \PackageError{active-bytes}{Missing final-figure readiness record}{Run the reviewed measured-data figure pipeline and write its digest before release.}%
  }
\fi

\newcommand{\TrafficRawRhoW}{0.9291}
\newcommand{\TrafficRawRhoKV}{1.0020}

\newcommand{\TrafficRawMAPE}{0.0498\%}

\newcommand{\TPtwoMedianError}{0.245\%}
\newcommand{\TPtwoMaximumError}{0.559\%}

\newcommand{\TPtwoPower}{314.10}

\newcommand{\GBQSevenMedian}{0.637\%}
\newcommand{\GBQSevenMaximum}{2.137\%}

\newcommand{\GBQFourteenHoldoutMedian}{0.172\%}
\newcommand{\GBQFourteenHoldoutMaximum}{0.971\%}

\newcommand{\GBMistralMedian}{0.565\%}
\newcommand{\GBMistralMaximum}{0.887\%}

\newcommand{\GBHBMRhoW}{0.928653}
\newcommand{\GBHBMRhoKV}{1.004953}

\newcommand{\GBHBMMaxCellDifference}{0.2384\%}

\providecommand{\GBHBMRhoKV}{}
\renewcommand{\GBHBMRhoKV}{1.004953}
\providecommand{\GBHBMRhoW}{}
\renewcommand{\GBHBMRhoW}{1.000784}

\providecommand{\TPtwoPower}{}
\renewcommand{\TPtwoPower}{314.101}
\providecommand{\TrafficRawRhoKV}{}
\renewcommand{\TrafficRawRhoKV}{1.002042}
\providecommand{\TrafficRawRhoW}{}
\renewcommand{\TrafficRawRhoW}{1.001213}

\newcommand{\Mw}{M_{\mathrm{w}}}
\newcommand{\Kkv}{K_{\mathrm{KV}}}
\newcommand{\Beff}{B_{\mathrm{eff}}}
\newcommand{\Lbar}{\bar L_{\mathrm{token}}}
\newcommand{\Atok}{A_{\mathrm{token}}}
\newcommand{\AtokRW}{A_{\mathrm{token}}^{\mathrm{rw}}}

\newcommand{\paperfigurebody}[4]{%
    \centering
    \IfFileExists{figures/#2}{%
      \includegraphics[width=#1]{figures/#2}%
    }{%
      \fbox{\parbox[c][2.5in][c]{0.88\linewidth}{%
        \centering\textbf{Figure placeholder}\\[0.5em]
        \texttt{\detokenize{figures/#2}}}}%
    }
    \caption{#3}
    \Description{#3}
    \label{#4}}

\title[The Token-Energy Law of LLM Decode]{\PaperTitle}

\author{Chenxu Niu}
\affiliation{%
  \institution{NVIDIA Corporation}
  \country{USA}}
\email{chenxun@nvidia.com}

\author{Batuhan Sencer}
\affiliation{%
  \institution{Texas Tech University}
  \country{USA}}
\email{batuhan.sencer@ttu.edu}

\author{Wei Zhang}
\affiliation{%
  \institution{Texas Advanced Computing Center}
  \country{USA}}
\email{wei.zhang@tacc.utexas.edu}

\author{Yong Chen}
\affiliation{%
  \institution{Texas Tech University}
  \country{USA}}
\email{yong.chen@ttu.edu}

\usepackage{wrapfig}

\begin{document}

\begin{abstract}
Energy-efficient deployment of large language models requires understanding how model architecture, hardware, and serving settings shape energy per generated token. We introduce the Token-Energy Law, a compact, interpretable model of GPU-board energy per output token for dense, full-history decoding. It relates token energy to two core dimensions: active bytes and realized decode time. Active bytes describe logical byte demand per output, separating dense weights shared within a decode iteration from private key--value (KV) histories read and updated for each request. Realized decode time is the elapsed decode duration divided by the number of output tokens produced. We combine the two byte coordinates and time in a common affine form, with coefficients calibrated for each model and hardware/runtime configuration. This formulation explains how batching amortizes shared-weight work, longer attended histories increase private KV demand, and throughput determines the time-associated cost per output. Our experiments cover three models under BF16 decoding on NVIDIA Hopper and Blackwell GPUs, including two-GPU tensor parallelism on Hopper. Using fixed coefficients and observed decode time, held-out condition-mean energy estimates differ from measurements by at most 3.73\%. Controlled request-departure tests further show that previously identified model-specific coefficients remain effective as batch membership changes, without recalibration for departure patterns. The central contribution is a reusable work--time relationship whose structure remains consistent across configurations while its coefficients describe local energy responses. It provides researchers and practitioners with a quantitative basis for interpreting inference energy and comparing serving trade-offs. Combined with independently validated performance models, the law can also support prospective energy estimation and energy-efficient deployment planning.

\end{abstract}

\keywords{large language model inference, energy measurement, scaling law,
 decode inference, KV cache, data movement}

\maketitle

\ifresultsready\else
  \PackageError{token-energy-law}{Measured results are required for this preprint}{}
\fi
\ifdurationevidencefinal\else
  \PackageError{token-energy-law}{Final duration evidence is required for this preprint}{}
\fi
\ifforecastresults
  \PackageError{token-energy-law}{Forecast mode is disabled for this preprint}{}
\fi

\section{Introduction}
\label{sec:introduction}

Large language models (LLMs) are increasingly deployed in applications ranging from information retrieval to code generation and conversational systems. 
The scale of this activity is growing rapidly. Google reports processing over 3.2 quadrillion tokens per month across its products in May 2026~\cite{pichai2026io}, compared with approximately 480 trillion a year earlier. At sufficient serving volume, cumulative inference energy can rival the energy invested in training; Luccioni et al. place this crossover at hundreds of millions of requests for the BLOOMz models they study~\cite{luccioni2024powerhungry}. Understanding the energy cost of inference is therefore an increasingly important problem in the design and operation of LLM services. This raises a fundamental systems question: \textbf{Can a common law explain how energy per generated token varies across models, hardware platforms, and deployment configurations?}

Research on inference energy spans benchmarking, workload-based estimation, and production measurement. TokenPowerBench~\cite{niu2025tokenpowerbench} and ML.ENERGY~\cite{chung2025mlenergy} characterize energy across models and serving configurations. Fernandez et al. examine the effects of workload geometry and execution configuration and develop measurement-based energy estimates for offline workloads~\cite{fernandez2025energy}. Google's production measurements account for the broader system costs of delivering AI services~\cite{elsworth2025googleenergy}. These studies provide an empirical foundation for understanding inference energy across models and serving configurations. Building on this foundation, we seek a common analytical law that relates energy per generated token to the underlying execution characteristics, providing a systematic explanation of how model and deployment choices affect energy consumption.

The difficulty of such an effort is that inference energy consumption depends on several interacting dimensions - \textbf{(1)} model architecture and numerical representation determine the weights and KV state used during generation;  \textbf{(2)}  hardware and kernels determine how that work is executed; \textbf{(3)} optimization strategies applied to deployment - such as batching, tensor parallelism, and pipeline parallelism - may affect work sharing, resource occupancy, communication, and execution time. Context length changes the history processed by attention. Nominal configuration parameters, however, do not always reflect the work performed during execution: quantization changes stored bytes without changing parameter count~\cite{lin2024awq}; iteration-level scheduling allows the active batch size to differ from the configured maximum~\cite{yu2022orca}; and allocated KV-cache capacity does not directly indicate the amount of history processed by attention~\cite{kwon2023pagedattention}. A useful energy model must identify execution quantities that capture these effects while retaining a consistent interpretation across configurations.

Recent analytical approaches already make progress toward this goal. SweetSpot~\cite{cavagna2026sweetspot} and From Tokens to Watt-hours~\cite{vartziotis2026tokens} relate inference energy to workload complexity and memory access; two frameworks named EnergyLens~\cite{palladino2026energylens,song2026energylens} develop closed-form and operator-level models for deployment choices. Request-level analyses also show why average token energy depends on the work included in the measurement window~\cite{vellaisamy2026requestenergy}. We address a specific question within this broader effort: \emph{can the sharing structure of decode supply a small set of interpretable work coordinates that, together with execution time, explain energy per useful output token across models and systems?} Our approach connects a trace-level accounting identity, an independently tested relationship to physical memory traffic, and a compact energy response.

Our starting point is exact decoding with dense weights and full attention history. Within an iteration, all useful outputs share the dense weight operators, while each output attends to its own KV history. To capture this sharing structure, we define \emph{active bytes} as the logical byte demand of shared dense-weight and private KV-cache accesses per useful output token. Let $M_w$ denote the shared dense-weight footprint in bytes, $\Beff$ the realized mean number of useful outputs per iteration, $\Kkv$ the combined key and value bytes per cached token (summed over layers), and $\Lbar$ the mean attended history length, weighted by useful tokens. The logical read demand per useful output from the weight and KV components is then $\Atok=\Mw/\Beff+\Kkv\Lbar$. The first term is shared work amortized by batch; the second is private work growing with context. Batch and context therefore affect per-token work in fundamentally different ways. Accounting for the write of one new KV entry per output gives the read--write demand $\AtokRW=\Mw/\Beff+\Kkv(\Lbar+1)$. These coordinates describe the trace's dense-weight and KV demand, not all resident memory or all operator accesses. Indexed embedding and positional lookups lie outside this principal account and require separate access accounting (Section~\ref{sec:background}); physical HBM measurements characterize how the
principal coordinates are realized by an implementation.

Energy depends on both active bytes and realized decode time. We express the weight and KV components in decimal GB as $q_{\mathrm w}=\Mw/(10^9\Beff)$ and $q_{\mathrm{KV}}=\Kkv(\Lbar+1)/10^9$, where the latter includes one new KV-record write per output. For decode duration $D$ and $T$ useful outputs, the realized time per token is $\tau=D/T$. The Token-Energy Law is 

\begin{equation}
\boxed{e_{\mathrm{gross},\sigma}=c_\sigma
+\alpha_\sigma q_{\mathrm w}+\beta_\sigma q_{\mathrm{KV}}
+P_{\tau,\sigma}\tau+u_\sigma .}
\label{eq:token-energy-law}
\end{equation}

Here $e_{\mathrm{gross},\sigma}$ is GPU-board energy in joules per useful output token. The coefficients $\alpha_\sigma$, $\beta_\sigma$, and $P_{\tau,\sigma}$ describe effective energy responses to weight demand, KV demand, and time; $c_\sigma$ is an approximately constant per-token overhead within execution setting $\sigma$, and $u_\sigma$ is the residual.

Generality lies in this reusable work--time structure, rather than universal energy-per-byte constants. We identify coefficients for each specified model and hardware--runtime setting $\sigma$ and test the empirical relationship on held-out workload configurations without refitting. The model provides an interpretable basis for comparing deployment choices under a common energy boundary. Our evaluation uses observed decode duration; prospective deployment estimates can combine the law with a separately validated duration model.

\paragraph{Contributions.}
\begin{itemize}
  \item \textbf{Trace-based active-byte accounting.} We derive active-byte coordinates from iteration-level weight sharing and token-weighted KV history. Their accounting accommodates changing request membership, and independent HBM measurements assess their correspondence with physical reads.

  \item \textbf{A unified work--time energy model.} We formulate a locally calibrated energy response over shared-weight demand, private-KV demand, and realized decode time. Frozen-coefficient evaluations test whether this response remains predictive across workload changes, including heterogeneous request departures.

  \item \textbf{Validation across models and execution settings.} Across three models on NVIDIA Hopper and Blackwell GPUs, our evaluation covers single-node tensor parallelism, same-SKU device transfer, and controlled request departures. All evaluated condition means from 374 held-out energy executions are within 3.73\% of measurement. Separately, 177 HBM profiles support the physical relevance of the byte coordinates, while matched GB300 ablations assess the contributions of realized time and separate weight/KV terms.
\end{itemize}

\section{Background and Related Work}
\label{sec:related}

Understanding inference energy requires connecting execution behavior to measured energy. We review the relevant serving mechanisms, measurement principles, and prior work that motivate our model.

\subsection{LLM Inference as a Serving Workload}
\label{sec:inference-background}

\paragraph{Prefill and autoregressive decode.}
During \emph{prefill}, the model processes the prompt and constructs the key and value (KV) cache. During conventional exact \emph{decode}, each active request generates one new token per iteration, using cached KV state to attend to its history. Prefill therefore exposes parallelism across input positions, whereas decode relies more heavily on concurrency across requests. Sarathi-Serve~\cite{agrawal2024sarathi} exploits this distinction by co-scheduling chunked prefills with ongoing decodes, whereas DistServe~\cite{zhong2024distserve} and Splitwise~\cite{patel2024splitwise} instead separate prefill and decode execution.

\paragraph{Batching and dynamic membership.}
Batching combines requests into a single execution that shares model weights across requests. It does not make their attention histories identical: each request still has its own sequence and attention work. Orca schedules at iteration granularity, allowing completed requests to leave and new requests to enter ~\cite{yu2022orca}. Consequently, a configured concurrency limit need not equal the number of useful outputs actually produced in each iteration. Similarly, context length is an evolving property of an active request, not just a maximum supported by the model. These distinctions matter when attributing work and energy to generated tokens rather than to a batch reservation.

\paragraph{Model representation and memory management.}
Parameter count alone does not determine the bytes accessed during inference. Weight precision changes storage and kernel behavior~\cite{lin2024awq,xiao2023smoothquant}; multi-query and grouped-query attention reduce the number of KV heads~\cite{shazeer2019multiquery,ainslie2023gqa}. KV quantization shrinks the per-token record, while history eviction alters the set of attended tokens~\cite{liu2024kivi,zhang2023h2o}. PagedAttention~\cite{kwon2023pagedattention} reduces allocation waste and enables KV sharing, and SGLang~\cite{zheng2024sglang} exploits reusable prefixes. Attention kernels can further reduce or reorganize transfers without changing the intended operator~\cite{dao2022flashattention,dao2024flashattention2,ye2025flashinfer}. Thus, allocated memory, logically attended state, and physical memory traffic are distinct quantities.

\paragraph{Hardware and parallel deployment.}
Accelerators differ in compute capability, memory bandwidth, on-chip storage, and execution efficiency. Tensor parallelism partitions work within layers and introduces collective communication; pipeline parallelism assigns layer groups to stages and introduces inter-stage dependencies and scheduling bubbles. Their costs depend on workload and mapping, rather than scaling directly with GPU count. LLMCompass and GenZ model such hardware and execution interactions~\cite{zhang2024llmcompass,bambhaniya2025genz}. An energy comparison must therefore specify both the deployment and the resources included in its measurement. For multi-GPU inference, energy comparisons should account for all participating GPUs.

\subsection{From Power Measurements to Token Energy}
\label{sec:energy-background}

Power is an instantaneous rate; energy accumulates over an execution interval. For measured power $P_{\mathcal E}(t)$ at boundary $\mathcal E$, an interval of duration $D$ and $T$ useful outputs has energy and average output-token energy:
\begin{equation}
E_{\mathcal E}=\int_0^D P_{\mathcal E}(t)\,dt,
\qquad e_{\mathcal E}=E_{\mathcal E}/T.
\label{eq:background-energy-accounting}
\end{equation}

A power trace, total request energy, and joules per output token answer different questions. Lower average power can coincide with higher energy if execution takes longer. Longer generation can increase request energy while reducing its average cost per token by amortizing request-level work, as Vellaisamy et al. demonstrate~\cite{vellaisamy2026requestenergy}.

The boundary also changes the interpretation. GPU-board energy, the sum over all GPUs in a replica, server energy, and datacenter energy include different components. TokenPowerBench~\cite{niu2025tokenpowerbench} separates inference phases and system-component contributions. Google's production study accounts for accelerators, host resources, provisioned idle capacity, and datacenter overhead~\cite{elsworth2025googleenergy}. Lifecycle and carbon models introduce further quantities, including embodied emissions and the carbon intensity of electricity~\cite{faiz2024llmcarbon,fu2025llmco2}. These are complementary accounting levels, not interchangeable energy measurements.

Token normalization additionally requires an explicit execution phase. Including prefill in the numerator charges prompt processing to the generated outputs; measuring decode alone isolates the recurring generation process. Speculative decoding performs draft and verification work in addition to accepted output generation~\cite{leviathan2023speculative}, while offloaded inference accesses resources beyond the GPU~\cite{sheng2023flexgen}. A common unit therefore requires consistent treatment of useful output, extra work, and participating resources. Our target is GPU-board energy of the declared decode execution, summed across participating GPUs when needed.

\subsection{What Measurement Studies Have Established}
\label{sec:measurement-related}

\paragraph{Cross-configuration characterization.} \emph{From Words to Watts} benchmarks LLaMA models of different sizes on V100 and A100 GPUs, using Alpaca and GSM8K workloads and model sharding across multiple GPUs and nodes~\cite{samsi2023wordswatts}. Its measurements relate runtime and energy to model size, batch size, output length, and hardware allocation. The study establishes that inference efficiency is a property of the complete configuration, not simply the parameter count of the model. Its central contribution is experimental characterization: it supplies measured operating trends rather than a shared-weight/private-KV token-energy equation derived from the execution trace.

\paragraph{Phase- and component-aware benchmarking.}
Niu et al.~\cite{niu2025energy} compare inference engines by separating setup and token-generation energy and measuring GPU, CPU, and DRAM contributions.
TokenPowerBench~\cite{niu2025tokenpowerbench} extends this measurement perspective by examining prefill and decode, model and precision choices, input/output lengths, batch size, and tensor- and pipeline-parallel configurations. It distinguishes GPU and other system-component contributions and connects observed behavior to utilization, memory access, and parallel execution. These decompositions motivate explanatory variables that relate deployment choices to energy while retaining phase and component boundaries.

\paragraph{Serving efficiency and energy diagnosis.}
ML.ENERGY~\cite{chung2025mlenergy} combines service-aware accounting with latency-constrained configuration selection, and relates energy efficiency to KV-cache memory, batch size, and low compute intensity of decoding. Subsequent work further diagnoses inference energy consumption~\cite{chung2026joules}. Fernandez et al.~\cite{fernandez2025energy} examine how model, workload, and execution choices affect inference energy and develop a measurement-based estimation method for binned offline workloads. They show that FLOP-based estimates assuming ideal hardware utilization can substantially underestimate energy consumption in the workloads they study. These studies thus include estimation and optimization as well as measurement. Luccioni et al.~\cite{luccioni2024powerhungry} compare energy across tasks and models; multi-request workflow studies extend the performance--energy question beyond isolated requests~\cite{ifath2026workflows}. Configuration sweeps expose coupled effects: batch size can change weight reuse, utilization, and duration, while context can change KV access, arithmetic work, and kernel choice. Explaining these trends requires identifying the work shared across outputs and the private work performed for each output.

\subsection{Analytical Models and the Level of Explanation}
\label{sec:analytical-related}

\begin{table}[tbp]
  \caption{Comparison of inference-energy measurement and modeling approaches.}
  \label{tab:related-work-matrix}
  \centering
  \small
  \setlength{\tabcolsep}{6pt}
  \renewcommand{\arraystretch}{1.0}
  \renewcommand{\tabularxcolumn}[1]{m{#1}}
  \begin{tabularx}{\linewidth}{@{}
    >{\centering\arraybackslash}m{0.27\linewidth}
    >{\raggedright\arraybackslash}X
    >{\raggedright\arraybackslash}X@{}}
    \toprule
    \textbf{Perspective and examples}
      & {\centering\bfseries Main question\par}
      & {\centering\bfseries Connection to this paper\par} \\
    \midrule

    Configuration measurement\par
      ~\cite{samsi2023wordswatts,niu2025tokenpowerbench,chung2025mlenergy}
      & Effects of model, workload, and serving choices on energy.
      & Execution variables that explain configuration-dependent
        trends. \\
        \midrule

    Analytical work--energy models~\cite{choi2013energyroofline,cavagna2026sweetspot,vartziotis2026tokens}
    \par
      & Energy costs of computation, data movement, and time.
      & Work--time foundation for shared-weight and KV
        accounting. \\
\midrule
    Deployment and operator\par predictors~\cite{palladino2026energylens,song2026energylens}
      & Energy prediction from configurations or operators.
      & Complements modeling based on realized trace quantities. \\
\midrule
    Request-window amortization~\cite{vellaisamy2026requestenergy}\par
      & Fixed and incremental energy within a request.
      & Explicit phase boundaries and useful-output normalization. \\

    \midrule
    \textbf{This work}\par
      \textbf{(Token-Energy Law)}
      & Decode energy from shared weights, private KV history,
        and realized time.
      & Trace-derived work coordinates, independent HBM measurements, and frozen-coefficient validation under static and changing request membership. \\
    \bottomrule
  \end{tabularx}
\end{table}

\paragraph{Work, data movement, and time.} Roofline relates performance to operational intensity and hardware limits~\cite{williams2009roofline}; Energy Roofline combines computation and data-movement costs with energy from constant power accumulated over execution time~\cite{choi2013energyroofline}. Hardware energy analyses and I/O lower bounds further connect representation, access location, and data movement to execution cost~\cite{horowitz2014energy,hong1981redblue,elango2015movementbounds,saha2024attentionio}. These foundations motivate work-based coordinates without equating logical demand with HBM traffic or assuming a universal energy cost per byte.

\paragraph{LLM-specific analytical and fitted predictors.} SweetSpot~\cite{cavagna2026sweetspot} derives efficiency relationships over input and output lengths from transformer compute and memory complexity. \emph{From Tokens to Watt-hours} separates computation, parameter
access, KV writes, and attention reads using hardware coefficients and calibrated reuse factors~\cite{vartziotis2026tokens}.
Palladino et al.'s EnergyLens~\cite{palladino2026energylens} uses symbolic regression to model energy across sequence lengths, batch sizes, and parallel configurations, with parameters fitted for different models and platforms.

\paragraph{Operator, hardware, and request models.} Song et al.'s EnergyLens~\cite{song2026energylens} combines operator and communication models, including execution overlap, for multi-GPU inference. GPUWattch, AccelWattch, and EnergAIzer model finer-grained GPU power~\cite{leng2013gpuwattch,kandiah2021accelwattch,lee2026energaizer}, while Vellaisamy et al. distinguish fixed request-window energy from incremental generation-step energy~\cite{vellaisamy2026requestenergy}. GenZ models distributed inference performance and estimates operator energy from execution time, static power, and utilization-dependent compute, memory, and network power~\cite{bambhaniya2025genz}. Table~\ref{tab:related-work-matrix} summarizes these complementary perspectives. Our question is whether iteration-level sharing and output-weighted history provide sufficient work coordinates for a compact energy response that remains predictive when request membership changes. We assess these coordinates independently through HBM measurements and test previously identified energy coefficients on controlled departure traces.

\subsection{Motivation}
\label{sec:law-motivation}

These studies motivate a decode-energy model organized around active bytes and realized time. The work coordinates should account for weights shared within an iteration and KV history processed by each output, including changes in request membership. Independent traffic measurements and held-out energy evaluations then test whether these quantities provide a useful basis for explaining energy across workloads.

\FloatBarrier
\section{The Law at a Glance}
\label{sec:law-glance}

The Token-Energy Law combines active bytes and realized time to describe energy per useful output.

\par\medskip
\noindent\begin{minipage}{\linewidth}
  \centering
  \small
  \textbf{Principal dense-weight and KV work}\quad
  $\displaystyle \AtokRW=\frac{\Mw}{\Beff}+\Kkv(\Lbar+1)$
  \par\medskip
  \begingroup
  \setlength{\fboxsep}{6pt}
  \colorbox{lawteal!8}{\begin{minipage}[c][0.78in][c]{0.28\linewidth}
    \centering\textcolor{lawteal}{\textbf{Shared dense weights}}\\[5pt]
    $\displaystyle q_{\mathrm w}=\frac{\Mw}{10^9\Beff}$\\[5pt]
    decimal GB / output token
  \end{minipage}}\hfill
  \colorbox{lawrose!8}{\begin{minipage}[c][0.78in][c]{0.28\linewidth}
    \centering\textcolor{lawrose}{\textbf{Private KV}}\\[5pt]
    $\displaystyle q_{\mathrm{KV}}=\frac{\Kkv(\Lbar+1)}{10^9}$\\[5pt]
    decimal GB / output token
  \end{minipage}}\hfill
  \colorbox{lawslate!9}{\begin{minipage}[c][0.78in][c]{0.28\linewidth}
    \centering\textcolor{lawslate}{\textbf{Realized time}}\\[5pt]
    $\displaystyle \tau=\frac{D}{T}$\\[5pt]
    seconds / output token
  \end{minipage}}
  \par\vspace{6pt}
  \begingroup
  \setlength{\fboxsep}{6pt}
  \setlength{\fboxrule}{0.8pt}
  \fcolorbox{lawteal!65!black}{lawteal!3}{%
    \begin{minipage}{\dimexpr\linewidth-2\fboxsep-2\fboxrule\relax}
      \centering
      {\normalsize\textbf{The Token-Energy Law}}\par\vspace{4pt}
      {\Large$\displaystyle
        \widehat e_{\mathrm{gross},\sigma}=c_\sigma
        +\textcolor{lawteal}{\alpha_\sigma q_{\mathrm w}}
        +\textcolor{lawrose}{\beta_\sigma q_{\mathrm{KV}}}
        +\textcolor{lawslate}{P_{\tau,\sigma}\tau}$}\par
    \end{minipage}}
  \endgroup
  \par\medskip
  \begin{minipage}{0.96\linewidth}
    \raggedright
    \textbf{Same form; stratum-specific coefficients.}
    $\sigma$ fixes the model, representation, hardware, runtime, topology,
    power policy, and energy boundary. The trace supplies the coordinates;
    local calibration supplies their energy response.
  \end{minipage}
  \endgroup
\end{minipage}
\par\medskip

\paragraph{Interpreting the coefficients.}
The active-byte coefficients $\alpha_\sigma$ and $\beta_\sigma$ express effective J/decimal-GB responses for shared-weight and private-KV demand. Their values can differ because these accesses have different reuse and access patterns. $P_{\tau,\sigma}$, in watts, describes time-associated energy conditional on the byte coordinates, while $c_\sigma$ represents an approximately workload-invariant per-token term within $\sigma$. These are fitted responses, not separately metered energy components; $u_\sigma$ is the remaining discrepancy.

\paragraph{Beyond $E=\bar P D$.}
The identity $E=\bar P D$ describes each run using its own average power. The law instead uses one locally calibrated coefficient set to estimate token energy across workload configurations from their active bytes and realized time. On the evaluated grid, retaining the weight and KV coordinates improves held-out agreement over a time-only model (Table~\ref{tab:gb300-ablation}).

\paragraph{Testable consequences.}
Within a fixed stratum, positive coefficients imply three conditional relationships. At fixed context and $\tau$, the shared-weight contribution decreases as $1/\Beff$. At fixed batch and $\tau$, the KV contribution grows linearly with attended history. At fixed active-byte coordinates, predicted energy increases linearly with $\tau$.

\FloatBarrier
\section{Formulating the Token-Energy Law}
\label{sec:background}

We formulate the Token-Energy Law using active bytes and realized decode time per useful output. We define the measurement boundary and useful-output count, then derive the shared-weight and private-KV coordinates and explain their roles alongside time in the energy response. 

\subsection{Defining the Token and the Energy Boundary}

Our unit is one \emph{output token} produced by dependency-ordered exact decode. The metered interval starts after an unmetered bootstrap makes the requests decode-ready and ends after the last counted output and device synchronization. Model loading and prefill are outside this interval. Let $T>0$ denote the number of output tokens generated within the metered decode interval and $D$ its duration in seconds. Each output token is counted once across the serving replica. Non-useful work consumes measured energy but does not increase the output count.

The energy boundary must match the serving replica whose work is counted. For a single-GPU replica, we use GPU-board energy. For a replica distributed across multiple GPUs, we sum board energies over the same synchronized interval and count each useful output once, not once per rank. With $\mathcal G$ denoting the replica's boards and $E_g$ the energy of board $g$, the energy outcome and time coordinate are
\begin{equation}
e_{\mathrm{gross}}=\frac{\sum_{g\in\mathcal G}E_g}{T}
\quad[\mathrm{J/token}],\qquad
\tau=\frac{D}{T}\quad[\mathrm{s/token}].
\label{eq:replica-energy-boundary}
\end{equation}
We report gross GPU-board energy without subtracting an idle baseline. The measurement covers the participating GPU boards, not the entire server or facility.

\begin{figure}[htbp]
  \centering
  \includegraphics[width=0.95\linewidth]
    {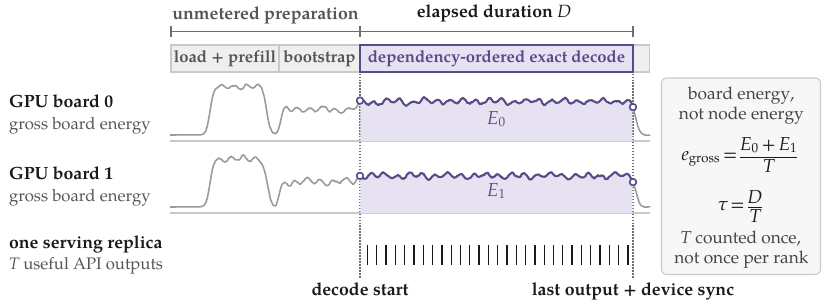}
  \caption{A common decode measurement interval for a two-GPU replica.
  Shaded areas under schematic power traces represent $E_0$ and $E_1$
  over duration $D$; $T$ counts useful outputs once for the replica.}
  \Description{Loading, prefill, and bootstrap precede metered decode.
  Two schematic GPU power traces share the same start and
  synchronization-completed end. Shaded areas represent each board's
  energy. A single row of output marks represents the replica-wide
  useful-output count. The right panel shows summed board energy
  and elapsed duration, each divided by that count.}
  \label{fig:measurement-boundary}
\end{figure}

Figure~\ref{fig:measurement-boundary} illustrates the shared measurement interval and replica-wide output count for two participating GPUs.

\paragraph{Realized batch and attended history.}
We characterize decode work using actual batch membership and attended history. For a trace with $S$ decode iterations, let $A_s$ denote the queries producing useful outputs in iteration $s$, and $L_{i,s}$ the number of historical records attended by query $i$ in that iteration. With $T=\sum_s |A_s|$, we define
\begin{equation}
\Beff=\frac{T}{S},\qquad
\Lbar=\frac{1}{T}\sum_{s=1}^{S}\sum_{i\in A_s}L_{i,s}.
\label{eq:effective-batch}
\end{equation}
Here $\Beff$ is the mean useful-output count per iteration, and $\Lbar$ is the mean attended history per useful output. Both describe executed work rather than configured concurrency or reserved KV capacity.

For $B$ synchronized requests, each attending $I$ historical records at the first metered step and producing $O$ outputs, we obtain $T=BO$, $\Beff=B$, and $\Lbar=I+(O-1)/2$, since each history grows by one record per iteration. When requests finish at different iterations, Equation~\ref{eq:effective-batch} accounts for their changing membership and histories. Section~\ref{sec:trace-validation} evaluates these coordinates in controlled request-departure experiments.

\subsection{Weight Work is Shared Across a Batch}

The weight component of active bytes follows from the sharing structure of dense decode. Within an iteration, the useful outputs share the same dense-weight operations. Let $\Mw$ denote the distinct weight bytes
required by these operations, assumed constant across the counted iterations. Over $S$ iterations producing $T$ useful outputs, the logical weight demand per output, expressed in decimal GB, is
\begin{equation}
q_{\mathrm w}
=\frac{S\Mw}{10^9T}
=\frac{\Mw}{10^9\Beff}.
\label{eq:weight-bytes}
\end{equation}
Doubling realized batch therefore halves the weight component of active bytes per output. It does not necessarily halve token energy, because private KV work and execution time need not decrease proportionally.
The count describes logical demand, not the number of times a kernel physically loads each byte from HBM.

$\Mw$ includes dense weights and associated parameter state consumed by the declared operations, rather than all resident model storage. Full input-embedding tables accessed through row lookups and positional lookup buffers are excluded; their accessed entries constitute auxiliary work. A tied embedding/output matrix is counted once when consumed by the dense output projection. For rank-resolved traffic accounting, aliased required storage is counted once within each rank, while distinct shards and physical replicas are summed across ranks. The TP=2 energy evaluation uses full-model reference coordinates rather than an audited rank-local storage sum (Section~\ref{sec:tp2-validation}).

Figure~\ref{fig:decode-sharing}(a,b) illustrates the distinction: increasing batch shares the same weight footprint across more outputs, while each output retains its own KV history. Panel~(c) plots the principal logical read--write demand $\Mw/B+\Kkv(L+1)$ for fixed history length $L$. Here $\Kkv$ is the size of a complete KV record, and the $+1$ accounts for one new-record write per useful output. As batch grows, weight demand falls and the combined demand approaches the KV term. Longer histories reach equal weight and KV demand at a smaller batch, given by $B^\star=\Mw/[\Kkv(L+1)]$. This is a balance of logical bytes, not an energy crossover. The next subsection derives the KV coordinate for a complete decode trace.

\begin{figure}[htbp]
  \centering
  \usetikzlibrary{arrows.meta,calc}
\begingroup
\long\def\TELFigureRender#1{\resizebox{\linewidth}{!}{#1}}
\definecolor{lawteal}{HTML}{247361}
\definecolor{lawrose}{HTML}{AF4C72}
\definecolor{lawslate}{HTML}{657F9B}
\definecolor{TELhistorylong}{HTML}{4B3C91}
\definecolor{TELhistoryshort}{HTML}{CB7B24}
\providecommand{\Mw}{M_{\mathrm{w}}}
\providecommand{\Kkv}{K_{\mathrm{KV}}}
\def\TELWeightGB{14.141238272}
\def\TELPlotLeft{9.65}
\def\TELPlotBottom{0.72}
\def\TELPlotWidth{4.30}
\def\TELPlotHeight{3.82}

\TELFigureRender{%
\begin{tikzpicture}[
  x=1cm,y=1cm,
  font=\sffamily\fontsize{8}{9.6}\selectfont,
  line cap=round,line join=round,
  TELtitle/.style={anchor=west,
    font=\sffamily\bfseries\fontsize{8.5}{10.2}\selectfont},
  weight/.style={draw=lawteal,fill=lawteal!10,line width=0.8pt,
    rounded corners=2pt,minimum width=3.35cm,minimum height=0.58cm},
  query/.style={circle,draw=lawslate,fill=lawslate!9,line width=0.7pt,
    minimum size=0.40cm,inner sep=0pt,
    font=\sffamily\fontsize{7}{8.4}\selectfont},
  history/.style={draw=lawrose,fill=lawrose!10,line width=0.7pt,
    minimum width=1.75cm,minimum height=0.44cm,inner sep=0pt},
  record/.style={draw=lawrose,fill=white,line width=0.7pt,
    minimum width=0.45cm,minimum height=0.44cm,inner sep=0pt,
    font=\sffamily\fontsize{7}{8.4}\selectfont},
  output/.style={circle,draw=lawslate,fill=lawslate!15,line width=0.7pt,
    minimum size=0.40cm,inner sep=0pt,
    font=\sffamily\fontsize{7}{8.4}\selectfont},
  flow/.style={-{Latex[length=1.1mm,width=0.8mm]},draw=lawslate!85!black,
    line width=0.6pt},
  share/.style={-{Latex[length=1.1mm,width=0.8mm]},draw=lawteal,
    line width=0.7pt}
]
  \path[use as bounding box] (0,-0.65) rectangle (14.10,5.42);
  \draw[lawslate!25,line width=0.5pt] (4.33,0.02)--(4.33,5.24);
  \draw[lawslate!25,line width=0.5pt] (8.77,0.02)--(8.77,5.24);

  \begin{scope}
    \node[TELtitle] at (0.10,5.10) {(a) One request: $B=1$};
    \node[weight] (wone) at (2.10,4.30)
      {Shared dense weights: $\Mw$};
    \node at (0.50,3.05) {query};
    \node at (2.10,3.05) {private KV};
    \node at (3.85,3.05) {output};

    \node[query] (qone) at (0.50,2.45) {$q_1$};
    \node[history] (hone) at (1.925,2.45) {KV$_1$: history $L$};
    \node[record] (none) at (3.025,2.45) {new};
    \node[output] (oone) at (3.85,2.45) {$y_1$};
    \draw[share] (wone.west)--(0.10,4.30)|-(qone.west);
    \draw[flow] (qone.east)--(hone.west);
    \draw[flow] (none.east)--(oone.west);

    \draw[lawslate!30,line width=0.5pt] (0.10,1.18)--(4.14,1.18);
    \node[anchor=west] at (0.18,0.89) {Per useful output};
    \node[anchor=west,text=lawteal!80!black] at (0.18,0.51)
      {Weight: $\Mw$};
    \node[anchor=west,text=lawrose!85!black] at (0.18,0.13)
      {KV: $\Kkv(L+1)$};
  \end{scope}

  \begin{scope}[xshift=4.45cm]
    \node[TELtitle] at (0.10,5.10) {(b) Three requests: $B=3$};
    \node[weight] (wthree) at (2.10,4.30)
      {Shared dense weights: $\Mw$};
    \node at (0.50,3.65) {query};
    \node at (2.10,3.65) {private KV};
    \node at (3.85,3.65) {output};

    \draw[lawteal,line width=0.7pt]
      (wthree.west)--(0.10,4.30)--(0.10,1.75);
    \foreach \i/\rowy in {1/3.15,2/2.45,3/1.75}{
      \node[query] (qthree\i) at (0.50,\rowy) {$q_{\i}$};
      \node[history] (hthree\i) at (1.925,\rowy)
        {KV$_{\i}$: history $L$};
      \node[record] (nthree\i) at (3.025,\rowy) {new};
      \node[output] (othree\i) at (3.85,\rowy) {$y_{\i}$};
      \draw[share] (0.10,\rowy)--(qthree\i.west);
      \draw[flow] (qthree\i.east)--(hthree\i.west);
      \draw[flow] (nthree\i.east)--(othree\i.west);
    }

    \draw[lawslate!30,line width=0.5pt] (0.10,1.18)--(4.14,1.18);
    \node[anchor=west] at (0.18,0.89) {Per useful output};
    \node[anchor=west,text=lawteal!80!black] at (0.18,0.51)
      {Weight: $\Mw/3$};
    \node[anchor=west,text=lawrose!85!black] at (0.18,0.13)
      {KV: $\Kkv(L+1)$};
  \end{scope}

  \node[anchor=north,align=center,text=black!65,
    font=\sffamily\fontsize{7}{8.4}\selectfont]
    at (4.34,-0.18)
    {Same model and history $L$; outlined boxes denote new KV records.};

  \begin{scope}[font=\sffamily\fontsize{7}{8.4}\selectfont]
    \node[TELtitle] at (9.00,5.10) {(c) Active bytes per output};
    \node[anchor=west,text=black!65] at (9.65,4.82) {Qwen2.5-7B, BF16};

    \foreach \Value/\Printed in {0.1/0.1,1/1,10/10}{
      \pgfmathsetmacro{\TickY}{\TELPlotBottom+
        \TELPlotHeight*ln(\Value/0.04)/ln(500)}
      \draw[lawslate!14,line width=0.3pt]
        (\TELPlotLeft,\TickY)--({\TELPlotLeft+\TELPlotWidth},\TickY);
      \draw[lawslate!85!black,line width=0.5pt]
        ({\TELPlotLeft-0.06},\TickY)--(\TELPlotLeft,\TickY);
      \node[anchor=east,inner sep=2pt]
        at ({\TELPlotLeft-0.06},\TickY) {$\Printed$};
    }
    \foreach \Power/\Printed in {0/1,2/4,4/16,6/64,8/256}{
      \pgfmathsetmacro{\TickX}{\TELPlotLeft+\TELPlotWidth*\Power/8}
      \draw[lawslate!85!black,line width=0.5pt]
        (\TickX,\TELPlotBottom)--(\TickX,{\TELPlotBottom-0.06});
      \node[anchor=north,inner sep=2pt]
        at (\TickX,{\TELPlotBottom-0.06}) {$\Printed$};
    }
    \draw[-{Latex[length=1.15mm,width=0.9mm]},lawslate!85!black,line width=0.6pt]
      (\TELPlotLeft,\TELPlotBottom)--
      ({\TELPlotLeft+\TELPlotWidth+0.05},\TELPlotBottom);
    \draw[-{Latex[length=1.15mm,width=0.9mm]},lawslate!85!black,line width=0.6pt]
      (\TELPlotLeft,\TELPlotBottom)--
      (\TELPlotLeft,{\TELPlotBottom+\TELPlotHeight+0.06});
    \node[rotate=90] at (8.98,2.65) {Logical demand (GB/output)};
    \node at (11.80,0.22) {Batch size $B$};

    \begin{scope}
      \clip (\TELPlotLeft,\TELPlotBottom)
        rectangle ({\TELPlotLeft+\TELPlotWidth},
                   {\TELPlotBottom+\TELPlotHeight});
      \draw[lawteal,line width=0.95pt]
        plot[domain=0:8,samples=121,variable=\t]
        ({\TELPlotLeft+\TELPlotWidth*\t/8},
         {\TELPlotBottom+\TELPlotHeight*
          ln((\TELWeightGB/pow(2,\t))/0.04)/ln(500)});

      \foreach \KVGB/\CurveColor in
        {0.234938368/TELhistoryshort,0.469819392/lawrose,0.939581440/TELhistorylong}{
        \pgfmathsetmacro{\FloorY}{\TELPlotBottom+
          \TELPlotHeight*ln(\KVGB/0.04)/ln(500)}
        \draw[\CurveColor,dash pattern=on 2pt off 1.6pt,line width=0.65pt]
          (\TELPlotLeft,\FloorY)--({\TELPlotLeft+\TELPlotWidth},\FloorY);
        \draw[\CurveColor,line width=1.05pt]
          plot[domain=0:8,samples=121,variable=\t]
          ({\TELPlotLeft+\TELPlotWidth*\t/8},
           {\TELPlotBottom+\TELPlotHeight*
            ln((\TELWeightGB/pow(2,\t)+\KVGB)/0.04)/ln(500)});
      }
    \end{scope}

    \node[anchor=west] at (10.52,4.32)
      {Total: $\Mw/B+\Kkv(L+1)$};
    \node[anchor=west,text=lawteal!85!black] at (10.84,3.96)
      {Weight only: $\Mw/B$};
    \node[anchor=west,font=\sffamily\fontsize{7}{8.4}\selectfont]
      at (11.40,3.48) {$B^*=\dfrac{\Mw}{\Kkv(L+1)}$};

    \foreach \KVGB/\Parity/\Rounded/\HistoryLabel/\CurveColor in {
        0.939581440/15.0505721697/15/16/TELhistorylong,
        0.469819392/30.0993073355/30/8/lawrose,
        0.234938368/60.1912680010/60/4/TELhistoryshort}{
      \pgfmathsetmacro{\FloorY}{\TELPlotBottom+
        \TELPlotHeight*ln(\KVGB/0.04)/ln(500)}
      \pgfmathsetmacro{\ParityX}{\TELPlotLeft+
        \TELPlotWidth*ln(\Parity)/ln(2)/8}
      \pgfmathsetmacro{\TotalY}{\TELPlotBottom+
        \TELPlotHeight*ln((2*\KVGB)/0.04)/ln(500)}
      \node[anchor=south west,text=\CurveColor,inner sep=1pt]
        at ({\TELPlotLeft+0.05},\FloorY) {$L=\HistoryLabel\mathrm{K}$};
      \draw[\CurveColor!55,densely dotted,line width=0.5pt]
        (\ParityX,\TELPlotBottom)--(\ParityX,\TotalY);
      \node[circle,draw=\CurveColor,fill=white,line width=0.9pt,
        minimum size=4.8pt,inner sep=0pt] at (\ParityX,\TotalY) {};
      \node[fill=white,inner sep=1pt,text=\CurveColor]
        at (\ParityX,0.97) {$\Rounded$};
    }
    \node[anchor=west,text=black!70] at (10.05,0.97) {$B^*\!\approx$};
    \node[anchor=north,align=center,text=black!65,
      font=\sffamily\fontsize{6.8}{8.2}\selectfont]
      at (11.74,-0.12)
      {Solid: total; dashed: KV; green: weights.\\
       Circles: equal weight/KV contributions; $\mathrm{K}=1024$.};
  \end{scope}
\end{tikzpicture}%
}
\endgroup
  \caption{Batch amortization of active bytes.
  (a,b) Weight demand is shared across outputs; KV histories remain private.
  (c) Analytical weight-plus-KV demand for Qwen2.5-7B (BF16).
  Circles mark equal weight and KV byte contributions.}
  \Description{Panels a and b compare weight sharing and private KV histories for batches of one and three. Panel c shows logical demand versus batch size for three history lengths.}
  \label{fig:decode-sharing}
\end{figure}

\subsection{KV Work Grows with Private History}

Unlike shared dense weights, each request processes its own private, materialized KV history. Let $\Kkv$ denote the constant byte size of one complete KV record across the declared serving replica. An output attending $L$ historical records incurs $\Kkv L$ logical read bytes and is aligned with one new-record write. Figure~\ref{fig:kv-active-bytes} illustrates this accounting over three consecutive decode steps.

Using the output-weighted mean history $\Lbar$ defined in Equation~\ref{eq:effective-batch}, the KV active-byte coordinate is
\begin{equation}
q_{\mathrm{KV}}=\frac{\Kkv(\Lbar+1)}{10^9}.
\label{eq:kv-bytes}
\end{equation}
The $+1$ counts one input-token KV write per useful output,
not an additional historical read. At fixed history and record
size, batching increases private KV work and useful outputs
proportionally, leaving the per-output coordinate unchanged.

\begin{figure}[htbp]
  \centering
  \usetikzlibrary{arrows.meta,calc}
\begingroup
\long\def\TELKVRender#1{\resizebox{\linewidth}{!}{#1}}
\definecolor{TELKVRose}{HTML}{AF4C72}
\definecolor{TELKVSlate}{HTML}{657F9B}
\definecolor{TELKVInk}{HTML}{243447}
\providecommand{\Kkv}{K_{\mathrm{KV}}}
\providecommand{\Lbar}{\overline{L}}

\TELKVRender{%
\begin{tikzpicture}[
  x=1cm,y=1cm,
  font=\sffamily\fontsize{8}{9.6}\selectfont,
  line cap=round,line join=round,
  title/.style={anchor=west,text=TELKVInk,
    font=\sffamily\bfseries\fontsize{8.8}{10.6}\selectfont},
  small/.style={text=TELKVInk!75,
    font=\sffamily\fontsize{7.5}{9}\selectfont},
  history/.style={draw=TELKVRose!75,fill=TELKVRose!19,
    line width=0.6pt,rounded corners=0.7pt,
    minimum width=0.53cm,minimum height=0.40cm,inner sep=0pt},
  newrecord/.style={draw=TELKVRose,fill=white,
    line width=0.95pt,rounded corners=0.7pt,
    minimum width=0.53cm,minimum height=0.40cm,inner sep=0pt},
  reuse/.style={-{Latex[length=1.2mm,width=0.9mm]},
    draw=TELKVRose!85,densely dashed,line width=0.6pt}
]
  \path[use as bounding box] (0,0) rectangle (14.65,4.02);

  \draw[TELKVSlate!25,line width=0.5pt] (7.75,0.12)--(7.75,3.95);

  \node[title] at (0.12,3.75) {(a) Read history, write one record};
  \node[anchor=west,small] at (0.12,3.34)
    {One request; three metered steps; initial history $L=4$};

  \node[small] at (0.54,2.93) {Step};
  \node[small] at (3.48,2.93) {Private KV records};
  \node[small] at (6.68,2.93) {Logical bytes};

  \foreach \Step/\Count/\RowY in {1/4/2.48,2/5/1.73,3/6/0.98}{
    \node[text=TELKVInk] at (0.54,\RowY) {$\Step$};
    \foreach \Record in {1,...,\Count}{
      \pgfmathsetmacro{\RecordX}{1.38+0.65*(\Record-1)}
      \node[history] (hist-\Step-\Record) at (\RecordX,\RowY) {};
    }
    \pgfmathsetmacro{\NewX}{1.38+0.65*\Count}
    \node[newrecord] (new-\Step) at (\NewX,\RowY) {};
    \node[text=TELKVInk] at (6.68,\RowY)
      {$(\Count+1)\Kkv$};
  }
  \draw[reuse] (new-1.south)--(hist-2-5.north);
  \draw[reuse] (new-2.south)--(hist-3-6.north);

  \node[history,minimum width=0.25cm,minimum height=0.19cm]
    at (0.30,0.38) {};
  \node[anchor=west,small] at (0.49,0.38) {History read};
  \node[newrecord,minimum width=0.25cm,minimum height=0.19cm]
    at (2.30,0.38) {};
  \node[anchor=west,small] at (2.50,0.38) {Input-token KV write};
  \node[anchor=west,small] at (5.22,0.38)
    {One block: $\Kkv$ bytes};

  \node[title] at (8.05,3.75) {(b) Normalize by useful outputs};
  \node[anchor=west,small] at (8.05,3.34)
    {Average history: $\Lbar=(4+5+6)/3=5$};

  \node[small] at (10.13,2.94) {History reads};
  \node[small] at (12.92,2.94) {Record writes};
  \node[text=TELKVRose!85!black,
    font=\sffamily\fontsize{10.5}{12.6}\selectfont]
    at (10.13,2.58) {$\Kkv(4+5+6)$};
  \node[text=TELKVInk] at (11.89,2.58) {$+$};
  \node[text=TELKVRose!85!black,
    font=\sffamily\fontsize{10.5}{12.6}\selectfont]
    at (12.92,2.58) {$3\Kkv$};
  \draw[TELKVSlate!70,line width=0.6pt] (8.83,2.27)--(14.08,2.27);
  \node[text=TELKVInk] at (11.45,2.03)
    {$3$ useful outputs};

  \draw[-{Latex[length=1.4mm,width=1mm]},TELKVSlate,
    line width=0.7pt] (11.45,1.80)--(11.45,1.51);

  \node[draw=TELKVRose!65,fill=TELKVRose!4,
    rounded corners=2pt,line width=0.7pt,
    minimum width=6.02cm,minimum height=1.08cm,inner sep=0pt]
    at (11.24,0.91) {};
  \node[text=TELKVInk,font=\sffamily\bfseries\fontsize{8}{9.6}\selectfont]
    at (11.24,1.20) {KV active bytes per useful output};
  \node[text=TELKVInk,font=\sffamily\fontsize{11.5}{13.8}\selectfont]
    at (11.24,0.72)
    {$\Kkv\Lbar+\Kkv=\Kkv(\Lbar+1)$};

  \node[anchor=west,small] at (8.05,0.12)
    {Logical read--write demand, not measured HBM traffic.};
\end{tikzpicture}%
}
\endgroup
  \caption{KV active bytes across three decode steps: growing
  history reads and one input-token KV write per useful output.}
  \Description{Three steps read four, five, and six historical
  records and each writes one input-token KV record. Normalizing
  by three useful outputs gives $\Kkv(\Lbar+1)$.}
  \label{fig:kv-active-bytes}
\end{figure}

For a packed, homogeneous, unreplicated layout, $\Kkv=2N_{\ell}N_{\mathrm{kv}}d_hs_{\mathrm{kv}}$, where the
factors count keys and values, layers, KV heads per layer, head dimension, and bytes per element. Multi-query and grouped-query attention reduce the KV-head count~\cite{shazeer2019multiquery,ainslie2023gqa}. Nonuniform or replicated layouts require summing the record footprints included in the declared accounting boundary.

\paragraph{Combining weight and KV active bytes.}
For constant $\Mw$ and $\Kkv$, let $D_{\mathrm{rw}}$ denote the total principal logical weight and KV read--write demand over the trace. Combining their contributions gives
\begin{equation}
\frac{D_{\mathrm{rw}}}{T}
=\AtokRW
=\frac{\Mw}{\Beff}+\Kkv(\Lbar+1).
\label{eq:active-byte-parity}
\end{equation}
This identity captures two distinct scaling effects: batching amortizes shared weights, whereas attended history determines private KV demand. These active bytes are computed from model and layout information and the realized decode trace, not inferred from measured HBM traffic. We next combine these work coordinates with realized time to model token energy.

\subsection{Modeling Token Energy with Active Bytes and Decode Time}
\label{sec:bound}

\begin{figure}[htbp]
  \centering
  \includegraphics[width=0.9\linewidth]{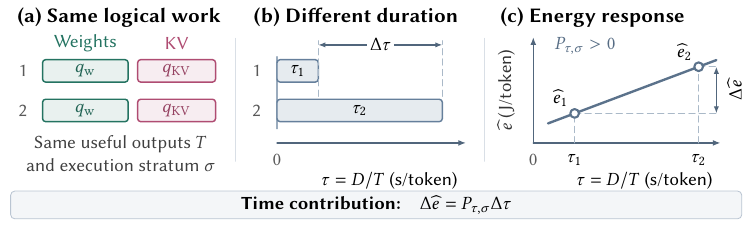}
  \caption{Conditional energy response to decode time at fixed active bytes, output count, and execution stratum ($P_{\tau,\sigma}>0$).}
  \Description{Three panels show equal logical work, different
  durations per output, and their conditional energy predictions.
  The predicted difference equals the time coefficient multiplied
  by the duration-per-output difference.}
  \label{fig:duration-response}
\end{figure}

Equal active-byte demand can accompany different execution times. Following the work--time perspective of energy roofline modeling~\cite{choi2013energyroofline}, we use $\tau=D/T$, the inverse aggregate useful-output throughput measured over the same interval. Within a fixed stratum $\sigma$, Figure~\ref{fig:duration-response} illustrates the conditional response $\Delta\widehat e=P_{\tau,\sigma}\Delta\tau$ at fixed active bytes and output count. For $P_{\tau,\sigma}>0$, longer duration raises predicted token energy. This effective coefficient has units of watts and is fitted jointly with the byte responses across workload cells; it is neither run-specific average power nor independently measured idle power.

We retain separate weight and KV coefficients because their reuse and kernel access patterns differ, allowing distinct energy responses per logical byte. For total dense-weight demand $S\Mw$, total KV demand $D_{\mathrm{KV}}=T\Kkv(\Lbar+1)$, and decode duration $D$, we propose the first-order additive response
\begin{equation}
E_{\sigma}=c_{\sigma}T
 +\alpha_{\sigma}\frac{S\Mw}{10^9}
 +\beta_{\sigma}\frac{D_{\mathrm{KV}}}{10^9}
 +P_{\tau,\sigma}D+U_{\sigma}.
\label{eq:extensive-energy-response}
\end{equation}
Dividing by the useful-output count $T$ yields the Token-Energy Law:
\begin{equation}
e_{\mathrm{gross},\sigma}=c_{\sigma}
 +\alpha_{\sigma}q_{\mathrm w}
 +\beta_{\sigma}q_{\mathrm{KV}}
 +P_{\tau,\sigma}\tau+u_{\sigma},
\label{eq:duration-augmented-law}
\end{equation}
where $u_{\sigma}=U_{\sigma}/T$.

The coefficients $\alpha_\sigma$ and $\beta_\sigma$, in J/decimal-GB, describe the effective responses to weight and KV demand. The intercept $c_\sigma$ captures an approximately constant per-output offset, and $u_\sigma$ is the remaining discrepancy. These are effective responses rather than separately metered physical components; auxiliary computation and runtime activity can influence both the coefficients and the residual. Within a fixed execution stratum, the byte footprints are constant, so the response is algebraically equivalent to a regression on inverse effective batch, token-weighted history, and time per output. The constant KV-append contribution cannot be identified separately from the intercept through this regression. The model uses one coefficient set within each declared stratum, with held-out workloads testing its adequacy across different batches, histories, and durations. A deployment change may affect several coordinates together, so their joint response matters. Our evaluation uses observed decode time; prediction before execution additionally requires an independently validated duration estimate.

Section~\ref{sec:evaluation} tests the active-byte coordinates against HBM measurements and evaluates the combined work--time response on held-out workloads and against simpler alternatives.

\FloatBarrier
\section{Evaluation and Validation}
\label{sec:evaluation}

We evaluate the Token-Energy Law using three dense language models on NVIDIA GH200 and GB300 GPUs. HBM traffic profiling examines how logical weight and KV demand correspond to physical memory traffic. Energy measurements assess the accuracy of predictions based on active bytes and decode time. Comparisons with simpler models establish the value of the separate weight, KV, and time terms, while experiments across devices, tensor-parallel configurations, and request-departure patterns test the law's applicability.

\subsection{Experimental Setup}

\subsubsection{Platforms and Workloads}

We study these three dense language models: Qwen2.5-7B-Instruct, Qwen2.5-14B-Instruct, and Mistral-7B-Instruct-v0.3~\cite{qwen2025qwen25,jiang2023mistral,mistral2024v03}. All experiments use BF16 weights and KV state, FlashAttention, and full-history attention. The GH200 platform is a dual-GPU node with 144-GB HBM3e per device, running ARM64 NVIDIA vLLM 25.09 with a fixed 700-W power limit per GPU. Single-GPU experiments use one device, while tensor-parallel experiments use both. The GB300 experiments use an ARM64 vLLM runtime validated for that platform, a fixed 1,300-W power limit per GPU, and tensor-parallel degree one. The evaluated vLLM V0 runtime uses hybrid CUDA-graph/eager execution with a default sequence-capture cap of 8192 tokens. Workload cells vary realized batch size and token-weighted attended history. Each static-batch cell is evaluated in five separate runs. After an unmetered bootstrap, each static decode episode generates 1,024 metered outputs per request on GH200 and 2,048 on GB300, with all requests remaining active throughout the episode. Each formal energy run accumulates at least 30 seconds of metered decode, summing energy, duration, and outputs across separately bracketed episodes and excluding intervening preparation. Separate GH200 trace experiments vary request output budgets and departure patterns, as described in Section~\ref{sec:trace-validation}. Iteration traces determine realized batch size and token-weighted attended history, while archived model inventories provide the weight and KV-layout information used to compute active bytes. Prefix caching, speculative decoding, chunked prefill, offload, and preemption are disabled.

\subsubsection{Energy and Traffic Measurement}

Active-byte coordinates are computed from archived BF16 model inventories and execution traces, independently of HBM measurements. The weight coordinate includes recurring dense operators and the output projection, but excludes the studied models’ untied input-embedding tables and rotary-position caches; recorded storage aliases are counted once. This inventory-based correction does not constitute a new kernel-access or rank-local audit. Corresponding coefficient rescaling preserves the separate-coordinate models’ predictions without changing the measurements. The corrected total-bytes baseline requires a post-hoc refit using the original training runs, as detailed in the ablation results.

The primary outcome is gross GPU-board energy per useful decode output. We integrate scope-0 instantaneous NVML power over synchronized decode boundaries, using the same interval for duration and useful-output count. Samples must bracket both endpoints, with no gap above 50 ms. On GH200, integrated scope-1 module power must also agree with the scope-matched cumulative energy counter within 2\%; GB300 uses the qualified scope-0 integral without this independent counter check. For two-GPU tensor parallelism, board energies are summed and normalized by the shared output count. Runs must satisfy the specified execution configuration, output counts, and scheduler constraints. Technical failures are retained but excluded from scientific repetitions; acceptance is independent of measured energy. Independent Nsight Compute executions measure HBM reads and writes for one post-bootstrap decode iteration. The dataset contains 177 profiles: 60 for Qwen7B, 45 for Qwen14B, and 45 for Mistral on GH200, plus 27 for Qwen7B on GB300. Application-range replay uses one metric range, one replay pass, zero warmup passes, and no profiler-controlled cache flushing or clock adjustment. The two-coordinate read model is fitted without an intercept, and its errors describe agreement on the profiled grid. Profiling is separate from all energy and latency measurements.

\subsubsection{Identification and Validation}

We estimate energy coefficients from run-level observations using ordinary least squares with HC3 standard errors~\cite{mackinnon1985hc3}. Fitting and held-out evaluation use disjoint batch--history cells. Separately interleaved executions are the statistical units. The primary GH200 Qwen7B response is fitted to 75 development observations, comprising all runs from the initial fitting, calibration, and evaluation stages. Its coefficients were frozen before collecting 45 new executions at nine previously unobserved batch--history cells. For the other six locally identified responses, residual-calibration observations do not enter the coefficient fits.

Table~\ref{tab:identification-protocol} lists the workload grids using $B=\Beff$ and $H=\Lbar/1024$. Each listed batch is paired with every listed history, with five runs per cell. Q7, Q14, and M7 denote Qwen2.5-7B, Qwen2.5-14B, and Mistral-7B; the four model--platform instances grouped in the second row are fitted separately.
\begin{table}[!htbp]
  \centering
  \caption{Workload grids for identification and held-out evaluation.}
  \label{tab:identification-protocol}
  \small
  \setlength{\tabcolsep}{4pt}
  \renewcommand{\arraystretch}{1.0}
  \renewcommand{\tabularxcolumn}[1]{m{#1}}
  \begin{tabularx}{\linewidth}{@{}
    >{\centering\arraybackslash}m{0.19\linewidth}
    >{\centering\arraybackslash}X
    >{\centering\arraybackslash}X
    >{\centering\arraybackslash}X@{}}
    \toprule
    Configuration
      & \shortstack{Fitting /\\development}
      & \shortstack{Residual\\calibration}
      & \shortstack{Held-out\\evaluation} \\
    \midrule
    GH200, Q7
      & $B\in\{4,8,16,24,32\}$\par $H\in\{4,8,16\}$\par 75 runs
      & No independent set
      & $B\in\{12,20,28\}$\par $H\in\{6,10,14\}$\par 45 runs \\
    \addlinespace[2pt]
    GH200 / GB300\par Q14 and M7
      & $B\in\{4,16\}$\par $H\in\{4,10,16\}$\par 30 runs each
      & $B=8$\par $H\in\{4,10,16\}$\par 15 runs each
      & $B\in\{6,12\}$\par $H\in\{6,12,14\}$\par 30 runs each \\
    \addlinespace[2pt]
    GH200, Q7\par TP=2
      & $B\in\{4,32\}$\par $H\in\{4,10,16\}$\par 30 runs
      & $B=16$\par $H\in\{4,10,16\}$\par 15 runs
      & $B\in\{8,24\}$\par $H\in\{6,12,14\}$\par 30 runs \\
    \addlinespace[2pt]
    GB300, Q7
      & $B\in\{4,32\}$\par $H\in\{4,8,16\}$\par 30 runs
      & $B=16$\par $H\in\{4,8,16\}$\par 15 runs
      & $B\in\{12,28\}$\par $H\in\{6,10,14\}$\par 30 runs \\
    \bottomrule
  \end{tabularx}
\end{table}
For each evaluation cell, we average measured and predicted energy per token separately and compute the absolute relative error as $|\bar e_j-\overline{\widehat e}_j|/\bar e_j$. Prespecified accuracy criteria require median and maximum cell-mean errors of at most 5\% and 10\%, respectively. Error bars show pointwise 95\% Student-$t$ confidence intervals for measured means; standard-deviation shading is identified separately in figure captions.

The same-SKU transfer test applies the GH200 GPU0 coefficients unchanged to 45 GPU1 executions on the same nine-cell evaluation grid. Static-batch validation therefore comprises 270 executions across 54 cohort-specific cells. Trace validation adds 40 executions each for Qwen7B and Mistral, followed by 24 additional Qwen repetitions reported separately. The resulting 374 validation executions exclude fitting, residual calibration, qualification, and HBM profiling.

\paragraph{Use of generative AI}
Generative AI tools, including OpenAI Codex, assisted with experimental planning, code development and debugging, data analysis and visualization, and manuscript drafting and revision. The authors take full responsibility for the methods, results, and conclusions presented in this paper.

\subsection{Active Bytes and Measured HBM Traffic}

\begin{figure}[!htbp]
  \centering
  \includegraphics[width=0.98\linewidth]{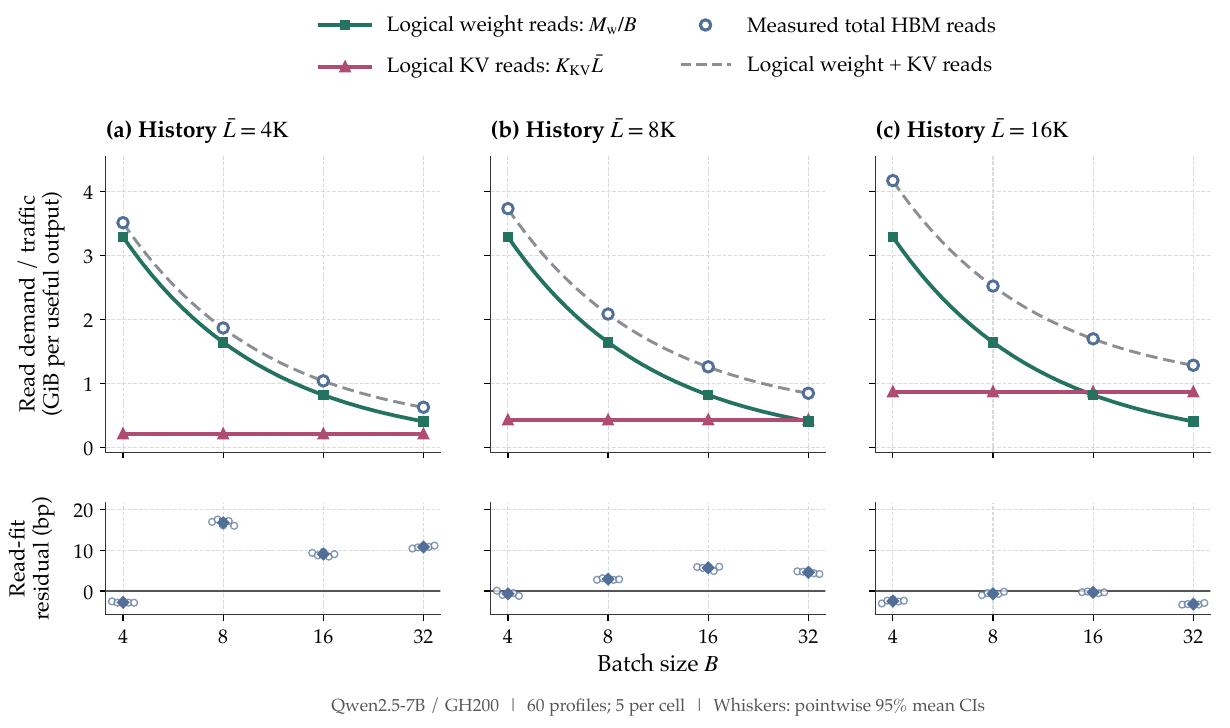}
  \caption{Qwen2.5-7B HBM reads on GH200. Top: logical weight/KV components, their unscaled sum, and measured means. Bottom: 60 fit residuals, $10^4(Q-\widehat Q)/\widehat Q$; diamonds mark five-run means. Whiskers show pointwise 95\% Student-$t$ confidence intervals.}
  \Description{Logical weight and KV demand compared with measured HBM reads, with fit residuals and confidence intervals for cell means.}
  \label{fig:raw-traffic-measured}
\end{figure}

The active-byte coordinates describe the logical byte demand of dense shared weights and private KV histories. Caching and kernel implementation affect how this demand translates into physical memory traffic. We therefore test whether the read coordinates capture measured HBM traffic using 177 profiles covering three models on GH200 and Qwen7B on GB300. Batch and history vary independently across the profiled grids. We fit the no-intercept model
\begin{equation} \frac{Q_{\mathrm{HBM,read}}^{\mathrm{obs}}}{T} = \rho_{\mathrm w}\frac{\Mw}{\Beff} + \rho_{\mathrm{KV}}\Kkv\Lbar + \varepsilon_{\mathrm{HBM}}, \label{eq:raw-traffic-fit} \end{equation}
where the dimensionless coefficients relate logical read demand to aggregate HBM reads.

\begin{table}[!htbp]
  \centering
  \caption{HBM-read fits on GH200 for $B\in\{4,8,16\}$ and $\bar L\in\{4096,8192,16384\}$ (45 profiles per model). Dense-sum gap is the mean absolute relative deviation of measured reads from logical weight-plus-KV reads, normalized by logical reads.}
  \label{tab:hbm-model-realization}
  \small
  \setlength{\tabcolsep}{4pt}
  \begin{tabular}{lrrrrr}
    \toprule
    Model & $\rho_{\mathrm w}$ & $\rho_{\mathrm{KV}}$ & MAPE & Max. APE & Dense-sum gap \\
    \midrule
    Qwen2.5-7B & 1.001189 & 1.002124 & 0.0460\% & 0.1757\% & 0.1714\% \\
    Qwen2.5-14B & 1.001661 & 1.000586 & 0.0324\% & 0.0822\% & 0.1494\% \\
    Mistral-7B & 1.001828 & 1.000494 & 0.0304\% & 0.0787\% & 0.1490\% \\
    \bottomrule
  \end{tabular}
\end{table}

For the full 60-profile Qwen7B grid on GH200, $\rho_{\mathrm w}=\TrafficRawRhoW$ and $\rho_{\mathrm{KV}}=\TrafficRawRhoKV$, with a mean absolute percentage error of \TrafficRawMAPE{} relative to measured reads. Figure~\ref{fig:raw-traffic-measured} shows the underlying work decomposition. Weight demand per useful output decreases with batch, while KV demand remains constant at fixed history and increases with history length. Measured HBM reads closely follow the unscaled sum of these logical demands; the lower panels show residuals relative to the fitted two-coordinate model.

Table~\ref{tab:hbm-model-realization} extends the comparison to three models on their shared nine-cell GH200 grid. For Qwen14B and Mistral, extending history from 4096 to 16384 at fixed batch increases measured reads by 0.99793--1.00045 times the additional logical KV demand, supporting the history-proportional KV term. The weight coefficients refer to the audited dense shared-weight footprint defined in the experimental setup.

\begin{wraptable}[11]{r}{0.48\textwidth}
  \centering
  \caption{Qwen2.5-7B HBM-read fits on matched grids.}
  \label{tab:gb300-hbm-realization}
  \small
  \setlength{\tabcolsep}{4pt}
  \begin{tabular}{@{}>{\centering\arraybackslash}m{0.4\linewidth}cc@{}}
    \toprule
    Metric & GH200 & GB300 \\
    \midrule
    Profiles & 45 & 27 \\
    $\rho_{\mathrm w}$ & 1.001003 & 1.000784 \\
    $\rho_{\mathrm{KV}}$ & 1.002258 & 1.004953 \\
    MAPE & 0.0444\% & 0.0730\% \\
    Max. APE & 0.1172\% & 0.2440\% \\
    \bottomrule
  \end{tabular}
\end{wraptable}

On GB300, the same form gives $\rho_{\mathrm w}=\GBHBMRhoW$ and $\rho_{\mathrm{KV}}=\GBHBMRhoKV$ (Table~\ref{tab:gb300-hbm-realization}). The matched cross-platform grid uses $B\in\{4,16,32\}$ and $\Lbar\in\{4096,8192,16384\}$. Across this grid, GH200 and GB300 cell-mean reads differ by at most \GBHBMMaxCellDifference{}. These results support the use of shared-weight and private-KV active bytes to describe HBM reads across the tested hardware and runtime configurations. These fits characterize aggregate reads on the profiled grids. Aggregate writes include traffic beyond logical KV appends and are outside this read-model validation.

\subsection{Validating the Work--Time Energy Model}

\begin{figure}[!htbp]
  \centering
  \includegraphics[width=0.98\linewidth]{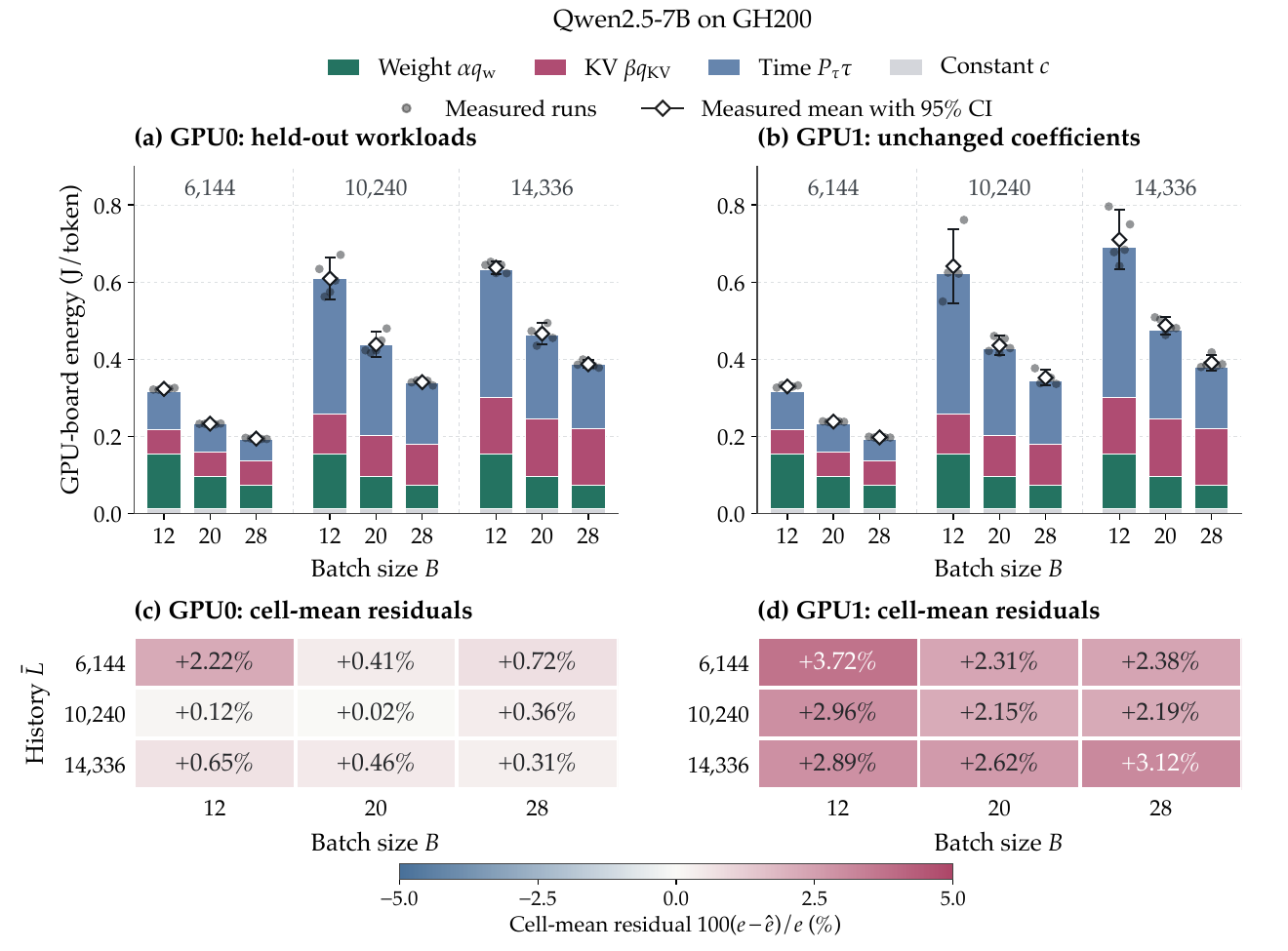}
  \caption{Qwen2.5-7B energy prediction on GH200 GPU0 and unchanged-coefficient transfer to GPU1. Stacks show model terms; diamonds and whiskers show measured means and 95\% confidence intervals. Residuals: $100(e-\widehat e)/e$.}
  \Description{Predicted and measured Qwen2.5-7B energy on GH200 GPU0 and GPU1, with model contributions, confidence intervals for measured means, and signed relative residuals.}
  \label{fig:duration-v2-validation}
\end{figure}

The Token-Energy Law predicts held-out decode energy using active-byte coordinates and observed decode time per useful output. For Qwen2.5-7B on GH200 GPU0, nine evaluation cells cross $\Lbar\in\{6144,10240,14336\}$ with $\Beff\in\{12,20,28\}$, with five independent executions per cell. With coefficients fixed before evaluation, the median and maximum absolute relative errors of cell means are \DurationVTwoMedianError{} and \DurationVTwoMaximumError{}, respectively. Figure~\ref{fig:duration-v2-validation}(a) compares the predicted energy contributions with measurements; panel (c) summarizes the cell-mean residuals.

The combined weight, KV, and observed-time terms track measured energy across the held-out workloads (Figure~\ref{fig:duration-v2-validation}). Stacked segments show fitted model contributions, not independently metered energy components. To assess the contribution of the modeling terms, Table~\ref{tab:gb300-ablation} compares four specifications for each GB300 model using the same 30 fitting runs and six held-out cells. These specifications retain separate weight and KV coordinates with time, separate byte coordinates alone, time alone, or a single total-byte coordinate with time. The total-bytes-plus-time specification constrains weight and KV demand to have the same energy coefficient.
\begin{table}[!htbp]
  \centering
  \caption{Maximum cell-mean absolute percentage error on GB300 holdouts (six cells and 30 runs per model).}
  \label{tab:gb300-ablation}
  \small
  \setlength{\tabcolsep}{4pt}
  \begin{tabular}{lrrrr}
    \toprule
    Model & Bytes + time & Bytes only & Time only & Total bytes + time \\
    \midrule
    Qwen2.5-7B & 2.137\% & 35.458\% & 29.395\% & 5.721\% \\
    Qwen2.5-14B & 0.971\% & 28.228\% & 10.972\% & 4.381\% \\
    Mistral-7B & 0.887\% & 33.564\% & 14.117\% & 5.589\% \\
    \bottomrule
  \end{tabular}
\end{table}

The complete work--time model gives the lowest maximum cell-mean error for all three models, ranging from 0.887\% to 2.137\%. Constraining weight and KV bytes to share one coefficient raises these errors to 4.381--5.721\%, although this specification also meets the prespecified median and maximum error thresholds. Separate weight and KV coefficients therefore improve accuracy on the tested grids. Omitting either realized time or the byte coordinates produces maximum errors above 10\% for all three models. The corrected total-bytes-plus-time comparison is a descriptive post-hoc reanalysis fitted only to the original training runs; held-out outcomes were not used to estimate its coefficients.

\subsection{Validation under Changing Request Membership}
\label{sec:trace-validation}

We apply the previously identified Qwen2.5-7B and Mistral-7B coefficients to GH200 decode traces in which requests finish at different iterations. Each model has eight conditions combining $\Beff\in\{10,20\}$ with a static control and three draining patterns. The draining patterns use uniform initial histories or positive or negative associations between initial history length and request lifetime. Drains begin with 16 or 32 requests that depart in four equal groups, with no new arrivals. All conditions match the token-weighted history $\Lbar=9983.5$. Five repetitions per condition give 40 runs per model; three additional Qwen repetitions per condition bring the total to 104 executions.

\begin{figure}[!htbp]
  \centering
  \includegraphics[width=0.98\linewidth]{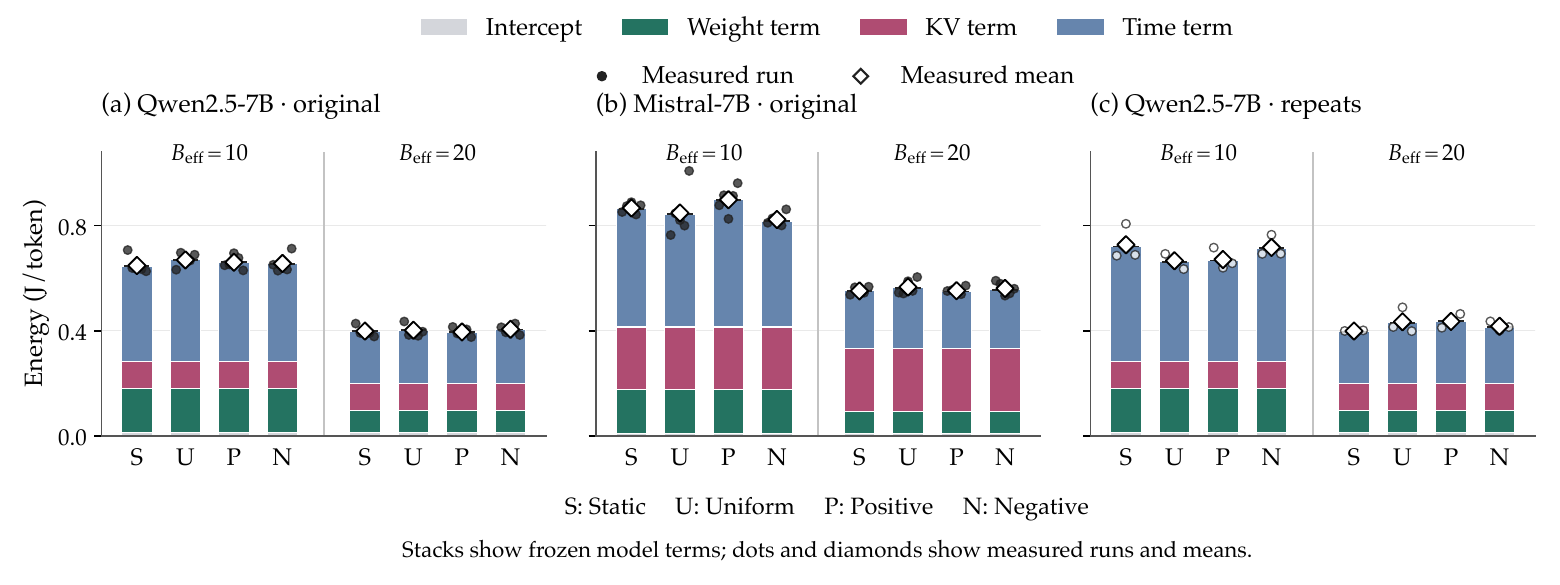}
  \caption{Frozen-coefficient predictions under changing request membership on GH200. Stacks show model terms; dots and diamonds denote measured runs and condition means.}
  \Description{Model contributions and measured decode energy for Qwen2.5-7B, Mistral-7B, and additional Qwen repetitions under changing request membership.}
  \label{fig:trace-energy-response}
\end{figure}

In Figure~\ref{fig:trace-energy-response}, matching $\Beff$ and $\Lbar$ within each model fixes the weight and KV contributions across departure patterns; differences in predicted energy enter only through observed $\tau$. Increasing $\Beff$ from 10 to 20 halves the weight term while leaving the matched-history KV term unchanged.

\begin{wraptable}{r}{0.48\textwidth}
  \centering
  \caption{Maximum condition-mean APE (\%) after coordinate substitution.}
  \label{tab:trace-coordinate-diagnostic}
  \small
  \setlength{\tabcolsep}{4pt}
  \begin{tabular}{@{}>{\centering\arraybackslash}m{0.38\linewidth}cc@{}}
    \toprule
    Coordinates & Q7 & M7 \\
    \midrule
    Realized & 0.404 & 0.870 \\
    $B_0$ replaces $\Beff$ & 9.824 & 8.258 \\
    $\bar L_{\mathrm{req}}$ replaces $\Lbar$ & 3.578 & 6.395 \\
    \bottomrule
  \end{tabular}
  
\end{wraptable}

We examine the coordinate definitions through post-hoc input substitutions on the original 40-run cohort for each model, excluding the 24 additional Qwen runs. Let $B_0$ denote initial cohort size and $\bar L_{\mathrm{req}}$ the equally weighted average of each request's mean attended history over its metered outputs. We replace either $\Beff$ with $B_0$ or $\Lbar$ with $\bar L_{\mathrm{req}}$, keeping the coefficients, observed duration, and all other inputs fixed.

Both substitutions increase the maximum condition-mean error despite retaining observed duration (Table~\ref{tab:trace-coordinate-diagnostic}). This diagnostic supports accounting for realized weight sharing and output-weighted KV history when transferring the fixed coefficients to changing request membership.

Every condition-mean error remains below 1\% within its evaluation cohort, including Mistral at $\Beff=20$, beyond its identification maximum of 16. Additional Qwen repetitions retain agreement across three blocks separated by at least two hours, supporting unchanged-coefficient transfer to these controlled departure patterns.

\subsection{Device Transfer and Tensor-Parallel Execution}
\label{sec:external}

\subsubsection{Transfer between physical devices}

The Qwen2.5-7B coefficients identified on GH200 GPU0 predict energy on GPU1 of the same node without refitting. The model, runtime, board power limit, and workload grid remain fixed. Across 45 executions at nine target-device cells, the maximum cell-mean error is 3.7234\%, rounded to 3.72\% (Figure~\ref{fig:duration-v2-validation}). Predicted cell means are consistently below the measured GPU1 means, while all cells satisfy the evaluation criterion. This result supports coefficient transfer between the two tested devices of the same SKU under the matched execution configuration.

\subsubsection{Two-GPU tensor-parallel decode}
\label{sec:tp2-validation}

The Token-Energy Law also describes Qwen2.5-7B decoding across two NVLink-connected GH200 GPUs using coefficients identified for this tensor-parallel configuration. Energy is summed across both boards and normalized by their common useful-output count. We evaluate the fixed coefficients on six held-out conditions crossing $\Lbar\in\{6144,12288,14336\}$ with $\Beff\in\{8,24\}$, with five executions per condition.

\begin{figure}[!htbp]
  \centering
  \includegraphics[width=0.98\linewidth]{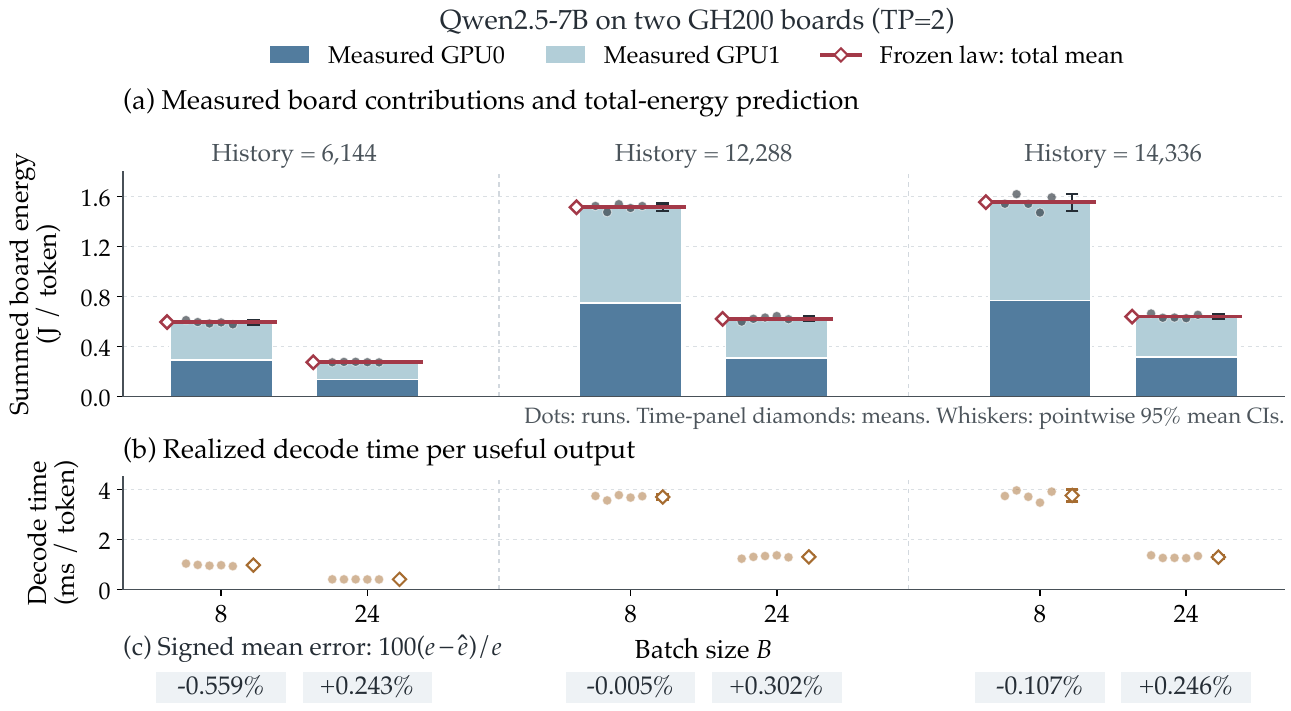}
  \caption{Qwen2.5-7B decoding on two GH200 boards. (a) Measured per-board energy and frozen total-energy predictions. (b) Realized decode time per useful output. (c) Signed errors of condition means. Dots show five runs per condition; whiskers are pointwise 95\% Student-$t$ confidence intervals for measured means.}
  \Description{Measured per-board energy and frozen summed-board predictions for two-GPU decode, alongside realized time per useful output and signed condition-mean errors.}
  \label{fig:tp2-validation}
\end{figure}

Figure~\ref{fig:tp2-validation} compares per-board measurements with frozen summed-board predictions, alongside observed time and signed errors. At each history, batch 24 has lower energy and shorter time per output than batch 8. Across six condition means, median and maximum absolute relative errors are \TPtwoMedianError{} and \TPtwoMaximumError{}, respectively.

These results support the same active-byte and time formulation for aggregate two-board energy after local calibration. The inputs use full-model reference byte footprints, and the fitted $P_\tau=\TPtwoPower$ W characterizes the deployment's time-associated energy response. Rank-local traffic and NVLink energy are not separately measured.

\subsection{Cross-Model and Cross-Platform Validation}
\label{sec:cross-platform-validation}
\label{sec:gb300}

The same active-byte and time formulation describes held-out decode energy for Qwen2.5-7B, Qwen2.5-14B, and Mistral-7B on GH200 and GB300. Qwen2.5-14B extends the evaluation to a larger model within the Qwen family, while Mistral-7B changes the architecture and KV geometry at a comparable parameter scale. Each model and execution configuration has its own coefficient vector, which remains fixed across the evaluated batches and histories.

\paragraph{Identified energy responses.}
The fitted instances make the common structure explicit. Let Q7, Q14, and M7 denote Qwen2.5-7B, Qwen2.5-14B, and Mistral-7B. Using the model-specific active-byte coordinates $q_{\mathrm w}$ and $q_{\mathrm{KV}}$ in decimal GB per useful output, and realized time $\tau$ in seconds per useful output, the single-GPU responses are
\begin{equation}
\begin{aligned}
\widehat e_{\mathrm{Q7}}^{\mathrm{GH200}}
&\simeq 0.0123 + 0.1194q_{\mathrm w} + 0.1801q_{\mathrm{KV}} + 139.56\tau,\\
\widehat e_{\mathrm{Q14}}^{\mathrm{GH200}}
&\simeq 0.0179 + 0.1190q_{\mathrm w} + 0.1837q_{\mathrm{KV}} + 139.07\tau,\\
\widehat e_{\mathrm{M7}}^{\mathrm{GH200}}
&\simeq 0.0095 + 0.1166q_{\mathrm w} + 0.1825q_{\mathrm{KV}} + 138.43\tau,
\end{aligned}
\label{eq:identified-gh200-laws}
\end{equation}
and
\begin{equation}
\begin{aligned}
\widehat e_{\mathrm{Q7}}^{\mathrm{GB300}}
&\simeq 0.0099 + 0.1173q_{\mathrm w} + 0.1915q_{\mathrm{KV}} + 198.57\tau,\\
\widehat e_{\mathrm{Q14}}^{\mathrm{GB300}}
&\simeq 0.0173 + 0.1163q_{\mathrm w} + 0.1927q_{\mathrm{KV}} + 199.33\tau,\\
\widehat e_{\mathrm{M7}}^{\mathrm{GB300}}
&\simeq 0.0108 + 0.1134q_{\mathrm w} + 0.1888q_{\mathrm{KV}} + 200.95\tau.
\end{aligned}
\label{eq:identified-gb300-laws}
\end{equation}
All responses give joules per useful output token. Coefficients are rounded for presentation; predictions use the full-precision values. For the two-GPU GH200 configuration evaluated in Section~\ref{sec:tp2-validation}, the separately identified response is
\begin{equation}
\widehat e_{\mathrm{Q7}}^{\mathrm{GH200,TP2}}
\simeq 0.0144 + 0.1204q_{\mathrm w} + 0.1806q_{\mathrm{KV}} + 314.10\tau.
\label{eq:identified-tp2-law}
\end{equation}
Here the outcome sums both boards' energy, and the byte coordinates use full-model reference footprints. The same-SKU device-transfer experiment instead retains the GH200 Q7 response unchanged.

Table~\ref{tab:coefficient-standard-errors} reports HC3 standard errors for the seven fitted responses. Across these fits, standardized design-matrix condition numbers range from 3.51 to 4.55, with a maximum variance-inflation factor of 5.69. The standard errors quantify coefficient uncertainty conditional on the observed design.

\begin{table}[!htbp]
  \centering
  \caption{HC3 standard errors of the fitted coefficients.}
  \label{tab:coefficient-standard-errors}
  \small
  \setlength{\tabcolsep}{4pt}
  \begin{tabular}{@{}cccccc@{}}
    \toprule
    Platform & Model & $\operatorname{SE}(c)$ & $\operatorname{SE}(\alpha)$ & $\operatorname{SE}(\beta)$ & $\operatorname{SE}(P_\tau)$ \\
    & & J/token & J/decimal-GB & J/decimal-GB & W \\
    \midrule
    GH200 & Q7  & 0.000861 & 0.000861 & 0.001947 & 0.820 \\
    GH200 & Q14 & 0.003848 & 0.000939 & 0.001830 & 0.986 \\
    GH200 & M7  & 0.002479 & 0.000876 & 0.001865 & 0.564 \\
    \midrule
    GB300 & Q7  & 0.001024 & 0.000431 & 0.002242 & 0.570 \\
    GB300 & Q14 & 0.002096 & 0.000359 & 0.001181 & 0.369 \\
    GB300 & M7  & 0.002688 & 0.001186 & 0.002215 & 0.959 \\
    \midrule
    GH200 TP=2 & Q7 & 0.001174 & 0.000434 & 0.002019 & 0.267 \\
    \bottomrule
  \end{tabular}
\end{table}

The coefficients characterize each model and hardware--runtime configuration; their time-associated energy contribution depends on both $P_\tau$ and the observed $\tau$.

\begin{table}[!htbp]
  \centering
  \caption{Static-workload energy validation. Errors compare predicted and measured cell means; each cell contains five executions. The device-transfer test retains the source-device coefficients.}
  \label{tab:energy-validation-summary}
  \small
  \setlength{\tabcolsep}{4pt}
  \begin{tabular}{lllrrrr}
    \toprule
    Platform & Model & Coefficients & Cells & Runs & Median & Maximum \\
    \midrule
    GH200 & Qwen2.5-7B & Local fit & 9 & 45 & \DurationVTwoMedianError & \DurationVTwoMaximumError \\
    GH200 & Qwen2.5-14B & Local fit & 6 & 30 & \QFourteenMedianError & \QFourteenMaximumError \\
    GH200 & Mistral-7B & Local fit & 6 & 30 & \MistralMedianError & \MistralMaximumError \\
    GH200 TP=2 & Qwen2.5-7B & Local fit & 6 & 30 & \TPtwoMedianError & \TPtwoMaximumError \\
    GB300 & Qwen2.5-7B & Local fit & 6 & 30 & \GBQSevenMedian & \GBQSevenMaximum \\
    GB300 & Qwen2.5-14B & Local fit & 6 & 30 & \GBQFourteenHoldoutMedian & \GBQFourteenHoldoutMaximum \\
    GB300 & Mistral-7B & Local fit & 6 & 30 & \GBMistralMedian & \GBMistralMaximum \\
    GH200 GPU1 & Qwen2.5-7B & Zero-refit transfer & 9 & 45 & \GPUOneTransferMedianError & \GPUOneTransferMaximumError \\
    \midrule
    \multicolumn{3}{l}{Static-batch validation subtotal} & 54 & 270 & --- & --- \\
    \bottomrule
  \end{tabular}
\end{table}

Table~\ref{tab:energy-validation-summary} summarizes the eight static-batch evaluations, comprising seven locally calibrated instances and one unchanged-coefficient device transfer. Across 270 held-out executions at 54 cohort-specific conditions, all condition-mean predictions are within 3.73\% of measurement.

\begin{figure}[!htbp]
  \centering
  \includegraphics[width=\linewidth]{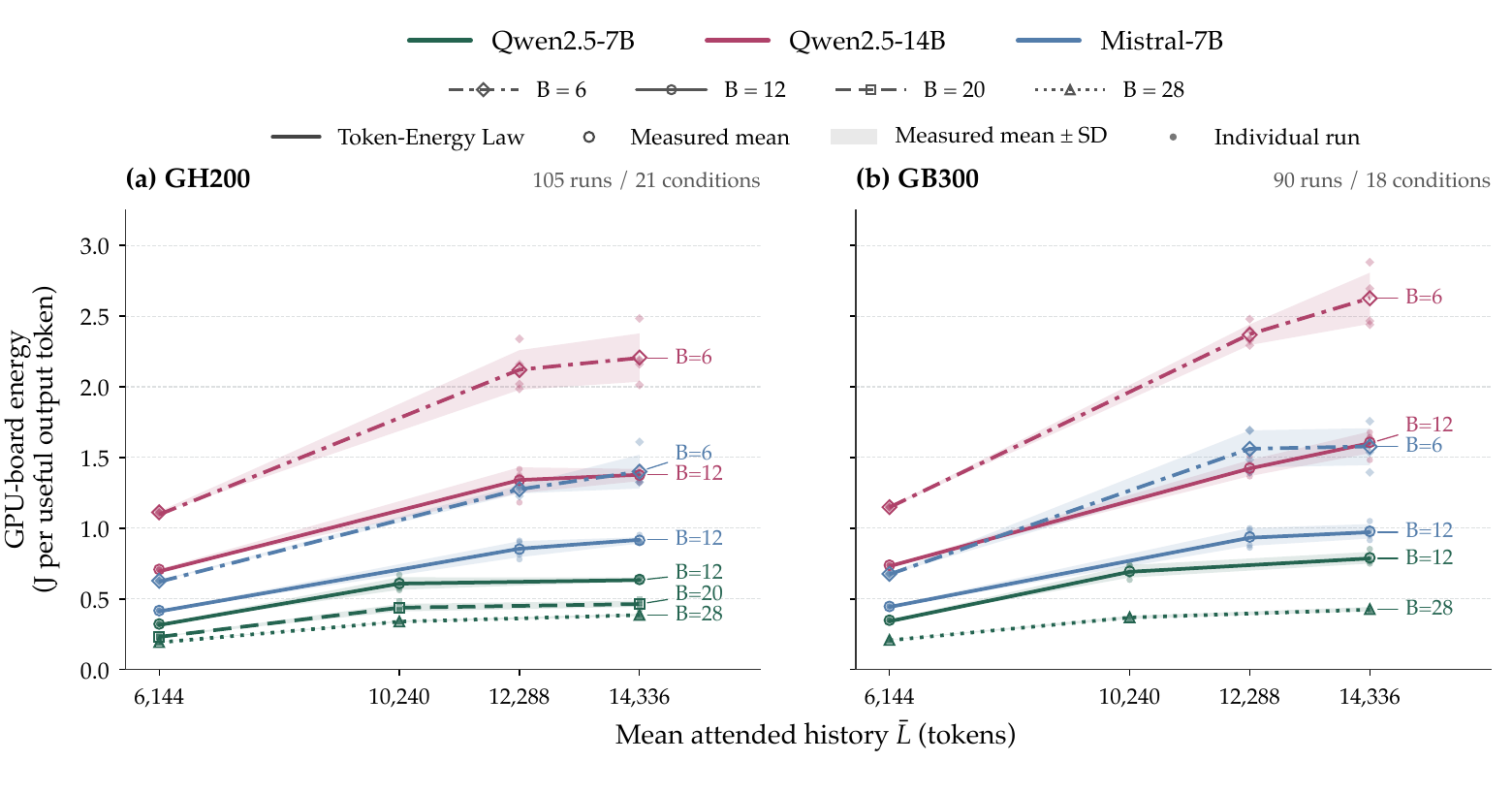}
  \caption{Held-out energy across models and platforms. Lines connect predictions using observed decode time; markers show measured means. Shading denotes $\pm1$ sample standard deviation across five runs per condition.}
  \Description{Two panels compare Token-Energy Law estimates with measured GPU-board energy per useful output on GH200 and GB300. The figure contains 195 executions at 39 conditions. Larger tested batches have lower measured mean energy at matched model, platform, and history settings. Mean energy increases with history in every plotted series. Shading represents repeated-measurement variability.}
  \label{fig:law-cross-platform}
\end{figure}

\paragraph{Held-out energy across models and platforms.} Figure~\ref{fig:law-cross-platform} summarizes 195 held-out executions at 39 conditions for six single-GPU model--platform pairs. Each pair retains one coefficient vector, with predictions evaluated using each run's active bytes and observed decode time. The figure shows two consistent workload effects across the tested models and platforms. At fixed model, platform, and history, the largest tested batch has 33.2--46.8\% lower measured mean energy per useful output than the smallest tested batch. In the law, batching amortizes shared-weight active bytes, while its accompanying execution-time changes enter through $\tau$. At fixed batch, longer histories increase private-KV active bytes, and measured mean energy rises in every plotted series. For GH200 Qwen2.5-7B at batch 12, mean energy rises from 0.323 to 0.609 to 0.638 J/token across histories 6144, 10240, and 14336. Archived step timings increase sharply when the maximum active sequence length exceeds 8192 tokens, consistent with eager fallback. These energy changes reflect both KV demand and runtime-dependent duration, which enters the predictions through observed $\tau$.

Across these six locally fitted single-GPU instances, the maximum condition-mean absolute relative error is 2.22\%. These results support the shared work--time form across the tested models and platforms without batch-specific or history-specific refitting.

\FloatBarrier
\section{Discussion and Limitations}
\label{sec:implications}

Our validation covers GPU-board energy for full-history dense BF16 decode under the evaluated runtime policy, excluding host-CPU energy. The maximum active sequence length exceeds the capture cap throughout the metered departure tests. Transfer to different graph policies, multi-node execution, and MoE requires further validation.

\paragraph{Multi-node and pipeline-parallel execution.}
Communication, synchronization, and pipeline bubbles change execution duration and device utilization. Although $\tau=D/T$ includes these delays, it does not resolve computation, communication, and waiting across individual GPUs. Equal active-byte demand and duration can therefore accompany different aggregate energy, and $P_\tau$ should not be interpreted as measured idle power. Our two-GPU NVLink experiments include communication effects within board energy, but do not validate multi-node or pipeline-parallel serving. Extensions require synchronized, boundary-matched measurements and may need stage- or communication-specific coordinates without double-counting existing terms. Network equipment outside the GPU-board boundary requires separate accounting.

\paragraph{Mixture-of-experts models.}
The coordinate $\Mw/\Beff$ assumes a fixed dense-weight footprint shared across outputs within each iteration. In routed MoE models, expert sharing depends on token assignments, so identical batch sizes and per-token activated parameter counts can imply different batch-level weight demands. Activated bytes cannot simply replace $\Mw$ and be divided by the full batch. Routing, placement, and caching affect physical traffic, while expert-parallel dispatch and load imbalance also affect duration. Extending the model requires routing-aware accounting of shared dense and expert work, followed by traffic and energy validation; recalibrating the dense coefficients alone is insufficient.

\section{Conclusion}

The Token-Energy Law provides a compact, interpretable account of GPU-board energy per useful output token for dense, full-history decoding. It combines two central factors: execution-derived active bytes and realized decode time. Active bytes distinguish dense-weight work shared across outputs from private KV work determined by attended history. Together with time per useful output, these coordinates yield a common energy response whose locally identified coefficients characterize the model and hardware--runtime configuration. Experiments across three dense models on Hopper and Blackwell support the same functional form with locally identified coefficients. Across the evaluated settings, held-out condition-mean energy estimates remain within 3.73\% of measurement.

The law gives researchers and system designers a quantitative basis for relating serving choices to token-level energy costs. Combined with independently validated performance models, it can support energy budgeting and comparisons among deployment options. Its explicit accounting of shared weights, private histories, and execution time also provides a testable starting point for extending energy models to other workloads and serving systems.

\bibliographystyle{ACM-Reference-Format}
\bibliography{references}

\end{document}